\documentclass[lettersize,journal,hidelinks]{IEEEtran}
\usepackage{amsmath,amsfonts}
\usepackage{algorithmic}
\usepackage{algorithm}
\usepackage{array}
\usepackage[caption=false,font=normalsize,labelfont=sf,textfont=sf]{subfig}
\usepackage{textcomp}
\usepackage{stfloats}
\usepackage{url}
\usepackage{verbatim}
\usepackage{graphicx}
\usepackage{cite}
\usepackage{pifont}
\newcommand{\cmark}{\ding{51}}
\newcommand{\xmark}{\ding{55}}
\usepackage{multirow}
\usepackage{arydshln}
\usepackage{hyperref}
\usepackage{orcidlink}
\usepackage{xcolor}

\begin{document}

\title{Dense Out-of-Distribution Detection \\ by Robust Learning  on Synthetic Negative Data}

\author{Matej Grcić\orcidlink{0000-0002-8379-6686
}, Petra Bevandić\orcidlink{0000-0002-8608-0624}, Zoran Kalafatić\orcidlink{0000-0001-8918-9070}, Siniša Šegvić\orcidlink{0000-0001-7378-0536}
\thanks{All authors are with the Faculty of Electrical Engineering and Computing, University of Zagreb, 10000 Zagreb, Croatia. Corresponding author e-mail: matej.grcic@fer.hr}
}

\markboth{Journal of \LaTeX\ Class Files,~Vol.~14, No.~8, August~2,21}%
{Shell \MakeLowercase{\textit{et al.}}: A Sample Article Using IEEEtran.cls for IEEE Journals}


\maketitle

\begin{abstract}
Standard machine learning is unable to accommodate inputs which do not belong to the training distribution. 
The resulting models often give rise to confident incorrect predictions which may lead to devastating consequences.
This problem is especially demanding in the context of dense prediction since input images may be only partially anomalous. Previous work has addressed dense out-of-distribution detection by discriminative training with respect to 
off-the-shelf negative datasets.
However, real negative data are unlikely to cover
all modes of the entire visual world.
To this end, we extend this approach by generating synthetic negative patches along the border of the inlier manifold.
We leverage a jointly trained normalizing flow due to coverage-oriented learning objective and the capability to generate samples at different resolutions. 
We detect anomalies according to a principled information-theoretic criterion which can be consistently applied through training and inference. 
The resulting models set the new state of the art on benchmarks for out-of-distribution detection in road-driving scenes and remote sensing imagery, in spite of minimal computational overhead.
\end{abstract}


\section{Introduction}
\IEEEPARstart{I}{mage} understanding involves recognizing objects 
and localizing them down to the pixel level 
\cite{everingham10ijcv}.
In its basic form, 
the task is to classify each pixel 
into one of K predefined classes,
which is also known as
semantic segmentation \cite{farabet13pami}.
Recent work improves perception quality
through instance recognition \cite{cheng20cvpr},
depth reconstruction \cite{godard19iccv},
semantic forecasting \cite{luc18eccv}, 
and competence in the open world \cite{uhlemeyer22uai}.

Modern semantic segmentation approaches \cite{farabet13pami} are based on deep learning.
A deep model for semantic segmentation maps the input RGB image $\mathbf{x}_{3\times H \times W}$ into the corresponding prediction $\mathbf{y}_{K\times H \times W}$.
Typically, the model parameters $\theta$ are obtained by gradient optimization of a supervised discriminative objective based on maximum likelihood.
Recent approaches produce high-fidelity segmentations of large images in real time even when inferring on a modest GPU \cite{orsic21pr}.
However, standard learning is susceptible to overconfidence in incorrect predictions \cite{guo17icml}, 
which may make the model unusable
in presence of semantic outliers \cite{zendel18eccv} and domain shift \cite{sakaridis21iccv}.
This poses a threat to models 
deployed in the real world \cite{chan21arxiv,blum21ijcv}.

We study ability of deep models for natural image understanding to deal with OOD input.
We desire to correctly segment the scene while simultaneously detecting anomalous objects which are unlike any scenery from the training dataset \cite{bevandic22ivc}.
Such capability is important in real-world applications like road driving \cite{lis19iccv,vojir21iccv} and remote sensing \cite{carvalho22rs,dasilva20arxiv}.

Previous approaches to dense OOD detection rely on Bayesian modeling \cite{kendall17nips}, image resynthesis \cite{biase21cvpr,lis19iccv,lis20arxiv}, recognition in the latent space \cite{blum21ijcv}, or
auxiliary negative training data \cite{bevandic19gcpr}. 
However, all these approaches have significant shortcomings.
Bayesian approaches and image resynthesis
require extraordinary computational resources
that hamper development and makes them 
unsuitable for real-time applications.
Recognition in the latent space \cite{blum21ijcv} may be sensitive to feature collapse \cite{amersfoort21arxiv,perera20cvpr}
due to relying on pre-trained features.
Training on auxiliary negative data may give rise to undesirable bias and over-optimistic evaluation.
Moreover, appropriate negative data may be unavailable in some application areas such as medical diagnostics \cite{gonzalez22mia} or remote sensing \cite{carvalho22rs,gawlikowski21igrass}.
Our experiments suggest that synthetic negatives may come to aid in such cases.

This work addresses dense out-of-distribution detection by encouraging the chosen standard dense prediction model to emit uniform predictions in outliers \cite{lee18iclr}. 
We propose to perform the training on mixed-content images \cite{bevandic19gcpr} which we craft by pasting synthetic negatives
into inlier training images.
We learn to generate synthetic negatives 
by jointly optimizing high inlier likelihood, 
and uniform discriminative prediction \cite{lee18iclr}.
We argue that normalizing flows  are better than GANs for the task at hand due to much better distribution coverage and more stable training.
Additionally, normalizing flows can generate samples of variable spatial dimensions \cite{dinh17iclr} which makes them suitable for mimicking anomalies of varying size.

This paper proposes five major improvements over our preliminary report \cite{grcic21visapp}.
First, we show that Jensen-Shannon divergence is 
a criterion of choice for robust joint learning 
in presence of noisy synthetic negatives.
We use the same criterion during inference,
as a score for OOD detection.
Second, we propose to discourage overfitting the discriminative model to synthetic outliers through separate pre-training of the discriminative model and the generative flow.
Third, we offer theoretical evidence for the advantage of our coverage-oriented synthetic negatives with respect to their adversarial counterparts.
Fourth, we demonstrate utility of synthetic outliers by performing experiments within the domain of remote sensing.
These experiments show that of-the-shelf negative datasets such as ImageNet, COCO or Ade20k do not represent a suitable source of negative content
for all possible domains.
Fifth, we show that training with synthetic negatives increases the separation between knowns and unknowns in the logit space, which makes our method a prominent component of future dense open-set recognition systems. 
We refer to the consolidated method as NFlowJS. 
NFlowJS achieves state-of-the-art performance on benchmarks for dense OOD detection in road driving scenes\cite{blum21ijcv,chan21arxiv} and remote sensing images \cite{carvalho22rs}, despite abstaining from auxiliary negative data \cite{bevandic19gcpr}, image resynthesis \cite{lis19iccv,biase21cvpr} and Bayesian modelling \cite{kendall17nips}.
Our method has a very low overhead over the standard discriminative model, making it suitable for real-time applications.

\section{Related work}

Several computer vision tasks require detection of unknown visual concepts
(Sec.\ \ref{sec:ood}).
In practice, this often has to be integrated with some primary classification task (Sec.\ \ref{sec:oodd} and \ref{sec:osr}).
Our method generates synthetic negatives with a normalizing flow
due to outstanding distribution coverage and 
capability to work at arbitrary resolutions (Sec.\ \ref{sec:gm}).

\subsection{Anomaly detection}
\label{sec:ood}

Anomaly detection, also known as novelty or out-of-distribution (OOD) detection, is a binary classification task which discriminates inliers from outliers \cite{hawkins80book,ruff21pieee}.
In-distribution samples, also known as inliers, are generated by the same generative process as the training data.
Contrary, anomalies are generated by a process which is disjoint from the training distribution \cite{zhang21icml}.
Samples of anomalous data may or may not be present during the training \cite{liang18iclr,hendrycks19iclr}.
The detection is typically carried out by thresholding some OOD score $s_\delta: [0, 1]^{3 \times H \times W} \rightarrow \mathcal{R}$ which assigns a scalar score to each test sample.
Some works address OOD detection in isolation, as a distinct computer vision task \cite{zhang21icml,zhou22tnnls,yang22tnnls,li14tpami,bergmann21ijcv,wang22tnnls,massoli22tnnls}.
Our work considers a different context where OOD detection is jointly solved with some discriminative task.

\subsection{Classification in presence of outliers}
\label{sec:oodd}

OOD detection \cite{hendrycks17iclr} can be implemented by extending  standard classifiers.
The resulting models can differentiate inliers while also detecting anomalous content.
A widely used baseline expresses the OOD score
directly from discriminative predictions as $s(x) = \max \mathrm{softmax}(\mathbf{f}_\theta(\mathbf{x}))$ \cite{hendrycks17iclr}.
Entropy-based detectors can deliver a similar performance \cite{macedo22tnnls,dhamija18nips}.
Another line of work improves upon these baselines 
by pre-processing the input with anti-adversarial perturbations \cite{liang18iclr}. 
Such perturbations cause significant 
computational overhead.

OOD detection has to deal with the fact that outliers and inliers may be indistinguishable in the feature space \cite{perera20cvpr}.
Feature collapse \cite{lucas19nips,amersfoort21arxiv} can be alleviated by training on negative data which can be sourced from real datasets \cite{hendrycks19iclr,dhamija18nips}
or generative models \cite{lee18iclr,grcic21visapp,du22iclr}.

There are two prior approaches
for replacing real negatives with synthetic ones \cite{lee18iclr,zhao23tpami}.
A seminal approach \cite{lee18iclr} proposes cooperative training
of a generative adversarial network and a standard classifier.
The classifier loss requires 
uniform predictions in generated samples
and thus encourages the generator
to yield samples at the distribution border.
This idea can be carried out without a separate generative model, by leveraging Langevin sampling \cite{zhao23tpami}.
However, adapting these approaches for dense prediction
is not straight-forward.
Similarly, synthetic outliers can be generated in the feature space by fitting GMM on known features \cite{du22iclr}.
However, our experiments indicate that this approach underperforms with respect to synthetic negative samples in input space.

Out-of-distribution detection gets even more complicated in the case of object detection and dense prediction where we have to deal with outlier objects in inlier scenes.
These models strive to detect unknown hazards while correctly recognizing the rest of the scene \cite{blum19iccvw,du22cvpr,riedlinger23wacv}.
A principled Bayesian approach to dense OOD detection attempts to estimate epistemic uncertainty \cite{kendall17nips}.
However, the assumption that MC dropout corresponds to Bayesian model sampling may not be satisfied in practice.
Another principled approach builds on 
likelihood estimation in feature space \cite{blum21ijcv}.
However, this may be vulnerable to feature collapse \cite{amersfoort21arxiv}.

Another line of work resynthesizes the input scene 
by processing dense predictions 
with a conditional generative model \cite{biase21cvpr,lis19iccv,xia20eccv}. 
Subsequently, anomalous pixels are detected
in reconstructive fashion \cite{ruff21pieee} 
by measuring dissimilarity between the input 
and the resyntesized image. 
Still, these approaches can detect anomalies only
in front of  simple backgrounds such as roads.
Also, resynthesis requires a significant computational budget which limits real-world applications.
A related approach utilizes a parallel upsampling path for input reconstruction \cite{vojir21iccv}.
This improves inference speed with respect to resynthesis approaches but still infers slower than our approach
while underperforming in cluttered scenes.

Several approaches train on mixed-content images
obtained by pasting negative patches into positive training examples \cite{bevandic19gcpr,biase21cvpr,chan21iccv}.
The negative dataset should be as broad as possible (eg. ImageNet or ADE20k) in order to cover a large portion of the background distribution.
The training can be implemented through a separate OOD head \cite{bevandic19gcpr} or by requiring uniform prediction in negative pixels \cite{chan21iccv}.
However, this kind of training results in biased models: test anomalies that are related to negative training data are going to give rise to above-average outlier detection performance.
Furthermore, competition on popular benchmarks may gradually adapt negative training data to test anomalies, and thus lead to over-optimistic performance estimates.
Our method avoids the bias of particular negative data
by crafting problem-specific negative samples
at the border of the inlier distribution.

\subsection{Open-set recognition}
\label{sec:osr}

Open-set recognition \cite{scheirer12tpami} discourages excessive generalization 
for known classes and attempts to distinguish them 
from the remaining visual content of the open world.
This goal can be achieved by rejecting classification in input samples which do not belong to the known taxonomy \cite{scheirer12tpami,bendale16cvpr,zhang20eccv,oliveira21ml}.
The rejection mechanism is usually implemented by restricting the shape of the decision boundary \cite{scheirer14tpami}.
This can be carried out by thresholding
the distance from learned class prototypes in the embedding
space \cite{cen21iccv,chen22tpami}.
Decision boundary can also be restricted by requiring a sufficiently large projection of the feature vector onto the closest class prototype \cite{hendrycks19arxiv}.
This is also known as max-logit detector which can be equally used for OOD detection and open-set recognition \cite{hendrycks19arxiv,vaze22iclr}.

Open-set recognition performance can be further improved by employing a stronger classifier \cite{vaze22iclr} or training on negative data \cite{neal18eccv,kong22tpami}.
Unlike OOD detection approaches based on softmax, open-set recognition methods provably bound open-space risk \cite{scheirer12tpami,boult19aaai}.
However, these approaches are still vulnerable to feature collapse \cite{amersfoort21arxiv}.
We direct the reader to \cite{geng21tpami,brilhadoz21} for a broader overview of open-set recognition.
Open-world approaches attempt to disentangle the detected unknown concepts towards new semantic classes.
This can be done in incremental \cite{michieli21cviu,uhlemeyer22uai} or low-shot \cite{yu18tnnls,shaban17bmvc,lu21tnnls} settings. 

Although we mainly focus on OOD detection, our synthetic negatives could be considered as synthetic known unknowns \cite{neal18eccv,kong22tpami}.
Our experimental evaluation suggests that our synthetic negatives increase the separation between known and unknown data in feature space.
This suggests that they may be helpful for open-set recognition \cite{hendrycks19arxiv,vaze22iclr}.



\begin{figure*}[!b]
    \centering
    \includegraphics[width=0.98\linewidth]{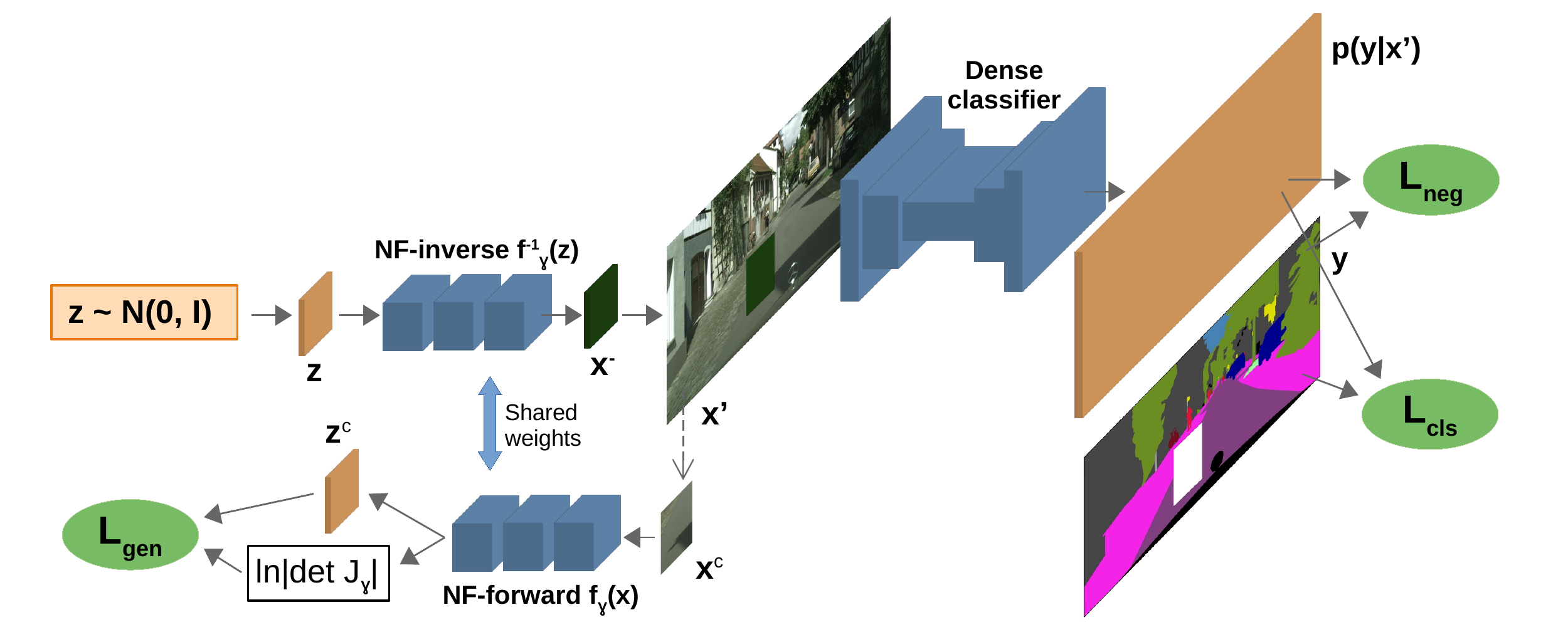}
    \caption{
    The proposed training setup.
    The normalizing flow generates the synthetic negative patch $\mathbf{x}^-$ which we paste atop the raw inlier image.
The resulting mixed-content image $\mathbf{x'}$ is fed to the dense classifier which is trained to discriminate inlier pixels ($L_{\mathrm{cls}}$) and to produce uniform predictions in negative pixels ($L_{\mathrm{neg}}$).
This formulation enables gradient flow from $L_{\mathrm{neg}}$ to the normalizing flow while maximizing the likelihood of inlier patches ($L_{\mathrm{gen}}$).
}
\label{fig:training-scheme}
\end{figure*}

\subsection{Generative models for synthetic negative data}
\label{sec:gm}
We briefly review generative approaches and
discuss their suitability for generating
synthetic negative training samples.
Energy-based \cite{salakhutdinov09aistats} and auto-regressive \cite{vanoord16icml} approaches are unsuitable for this task due to slow sampling.
Gaussian mixtures are capable of generating synthetic samples in the feature space \cite{du22iclr}.
VAEs \cite{kingma14iclr} struggle with unstable training \cite{vahdat20neurips} and have to store both the encoder and the decoder in GPU memory. 
GANs \cite{goodfellow20acm} also require lots of GPU memory and the produced samples do not span the entire support of the training distribution \cite{lucas19nips}.
On the contrary, normalizing flows \cite{dinh17iclr} offer efficient sampling and outstanding distribution coverage \cite{grcic21neurips}.

Normalizing flows \cite{dinh17iclr,kingma18neurips} model the likelihood as bijective mapping towards a predefined latent distribution $p(\mathbf{z})$, typically a fully factorized Gaussian.
Given a diffeomorphism $f_\gamma$, the likelihood is defined according to the change of variables formula:
\begin{equation}
\label{eq:cov}
    p_\gamma(\mathbf{x}) = p(\mathbf{z}) \left|\det \frac{\partial \mathbf{z}}{\partial \mathbf{x}}\right|, \quad \mathbf{z} = f_\gamma(\mathbf{x}).
\end{equation}
This setup can be further improved by introducing stochastic skip connections which increase the efficiency of training and improve convergence speed \cite{grcic21neurips}.

A normalizing flow $f_\gamma$ can be sampled in two steps.
First, we sample the latent distribution to obtain the factorized latent tensor $\mathbf{z}$.
Second, we recover the corresponding image through the inverse transformation $\mathbf{x} = f_\gamma^{-1}(\mathbf{z})$.
Both the latent representation and the generated image have the same dimensionality ($ \mathcal{R}^{3\times H \times W} \rightarrow [0, 1]^{3\times H \times W}$).
This property is useful for generating synthetic negatives since it allows to sample the same model on different spatial resolutions \cite{dinh17iclr}.

\section{Dense OOD detection with NFlowJS}

We train dense OOD detection on
mixed-content images
obtained by pasting synthetic negatives
into regular training images.
We generate such negatives
by a jointly trained normalizing flow (Sec.\ \ref{sec:neg_ds_to_flow}).
We train our models to recognize outliers
according to a robust information-theoretic criterion (Sec.\ \ref{sec:div-loss}),
and use the same criterion as our OOD score
during inference (Sec.\ \ref{sec:div-inf}).
Finally, we present a theoretical analysis which advocates for training with synthetic negatives generated through likelihood maximisation (Sec.\ \ref{sec:theoretical_analysis}).
\subsection{Training with synthetic negative data}
\label{sec:neg_ds_to_flow}

We assemble a mixed-content image $\mathbf{x}'$
by sampling a randomly sized negative patch $\mathbf{x}^-$ 
from a jointly trained normalizing flow $f_\gamma$,
and pasting it atop the inlier image $\mathbf{x}^+$:
\begin{equation}
    \mathbf{x}' = (\mathbf{1}-\mathbf{s}) \cdot \mathbf{x}^+ +  \mathrm{pad}(\mathbf{x}^-,\mathbf{s}) \; \mathrm{where} \; \mathbf{x}^- = f_{\gamma}^{-1}(\mathbf{z}).
\end{equation}
The binary mask $\mathbf{s}$ identifies pixels
of a pasted synthetic negative patch
within the input mixed-content image.
As usual in normalizing flows, $\mathbf{z}$ is sampled from a factorized Gaussian and reshaped according to desired spatial resolution.
The negative patch $\mathbf{x}^-$ is zero-padded in order to allow pasting by addition.
The pasting location is selected randomly.

We train our discriminative model by minimizing cross-entropy over inliers $(\textbf{s}^{ij}=0)$  and 
maximizing prediction entropy in pasted negatives $(\textbf{s}^{ij}=1) $\cite{hendrycks19iclr,lee18iclr,dhamija18nips}:
\begin{equation}
\label{eq:loss_seg}
    L_\text{disc}(\theta;\gamma) = \sum_{i, j}^{H,W} (\mathbf{s}^{ij} - 1)\cdot \ln p_\theta(\mathbf{y}^{ij}|\mathbf{x}') + \lambda \cdot  \mathbf{s}^{ij} \cdot L_{\mathrm{neg}}^{ij}(\theta,\gamma).
\end{equation}

We jointly train the normalizing flow alongside the primary discriminative model (cf.\ Figure \ref{fig:training-scheme})
in order to satisfy two opposing criteria.
First, the normalizing flow should maximize the likelihood of inlier patches
Second, the discriminative model should yield uniform distribution
in generated pixels. 
The former criterion aligns the generative distribution
with the inliers, while the latter pulls them apart. 
Such training encourages generation of synthetic samples at the boundary of the training distribution and incorporates outlier awareness within the primary discriminative model \cite{lee18iclr}.
The total loss applied to the generative model equals to:
\begin{equation}
\label{eq:loss_joint}
    L_\text{gen}(\gamma;\theta) = L_{\mathrm{nll}}(\gamma) + \lambda \sum_{i, j}^{H,W} \mathbf{s}^{ij} \cdot  L_{\mathrm{neg}}^{ij}(\theta, \gamma).
\end{equation}
$L_{\mathrm{nll}}$ is the negative log-likelihood of the inlier patch which gets replaced with the synthetic sample.
Formally, we have $L_\text{nll}(\gamma) = - \ln p_\gamma(\mathbf{x}^c)$
where $p_\gamma$ is defined in (\ref{eq:cov}).
We scrutinize $L_{\mathrm{neg}}$ in the following section.
The end-to-end training procedure minimizes the following loss: 
\begin{equation}
\label{eq:joi_e2e}
    L(\theta,\gamma) = L_\text{disc}(\theta;\gamma) + L_\text{gen}(\gamma;\theta).
\end{equation}

\begin{figure}[ht]
    \centering
    \includegraphics[width=\linewidth]{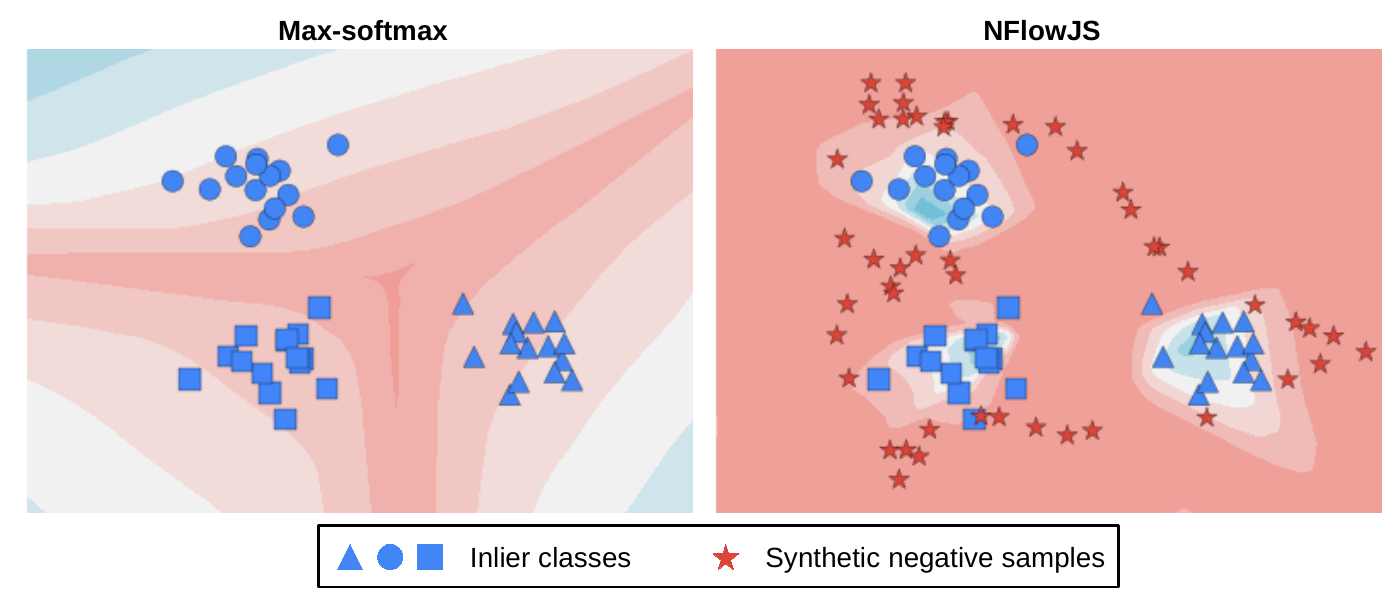}
    \caption{Softmax-activated discriminative models
do not bound the input-space volume
with confident predictions 
(blue region, left).
We address this issue by learning
a generative normalizing flow
for a "negative" distribution 
that encompasses the training manifold
(red stars, right).
Training the discriminative model 
to predict high entropy 
in the generated 
synthetic negative samples
decreases the confidence
outside the inlier manifold 
(red region, right).}
    \label{fig:toy_nfjs}
\end{figure}

Given enough training data and appropriate capacity, our synthetic negatives are going to encompass the inlier manifold.
Consequently, our method stands fair chance to detect visual anomalies that had not been seen during training due to being closer to synthetic negatives than to the inliers.
Figure \ref{fig:toy_nfjs} shows this on a 2D toy example.
The red color corresponds to higher values of the OOD score.
The left plot presents the max-softmax baseline \cite{hendrycks17iclr} which assigns high OOD score only at the border between the inlier classes.
The right plot corresponds to our setup which discourages low OOD scores 
outside the inlier manifold.
Synthetic negatives are denoted with red stars, while inlier classes are colored in blue.

\subsection{Loss in synthetic negative pixels}
\label{sec:div-loss}

The loss $L_{\mathrm{neg}}$ has often been designed as KL-divergence between the uniform distribution and the model's predictive distribution \cite{lee18iclr,hendrycks17iclr,dhamija18nips}.
However, our generative model is also subjected to the $L_\text{nll}$ loss.
Hence, the generated samples occasionally contain parts 
 very similar to chunks of inlier scenes, which lead to confident predictions into a known class.
Unfortunately, such predictions lead to unbounded penalization by KL divergence and can disturb the classifier which is also affected by $L_\text{neg}$.
If $L_\text{neg}$ overrides $L_\text{disc}$ in such pixels, then the classifier may assign high uncertainty in inliers. 
In that case, 
the incidence of false positive anomalies would severely increase. 
We address this problem by searching 
for a more robust formulation of $L_\text{neg}$.

\begin{figure}[!ht]
    \centering
    \includegraphics[width=\linewidth]{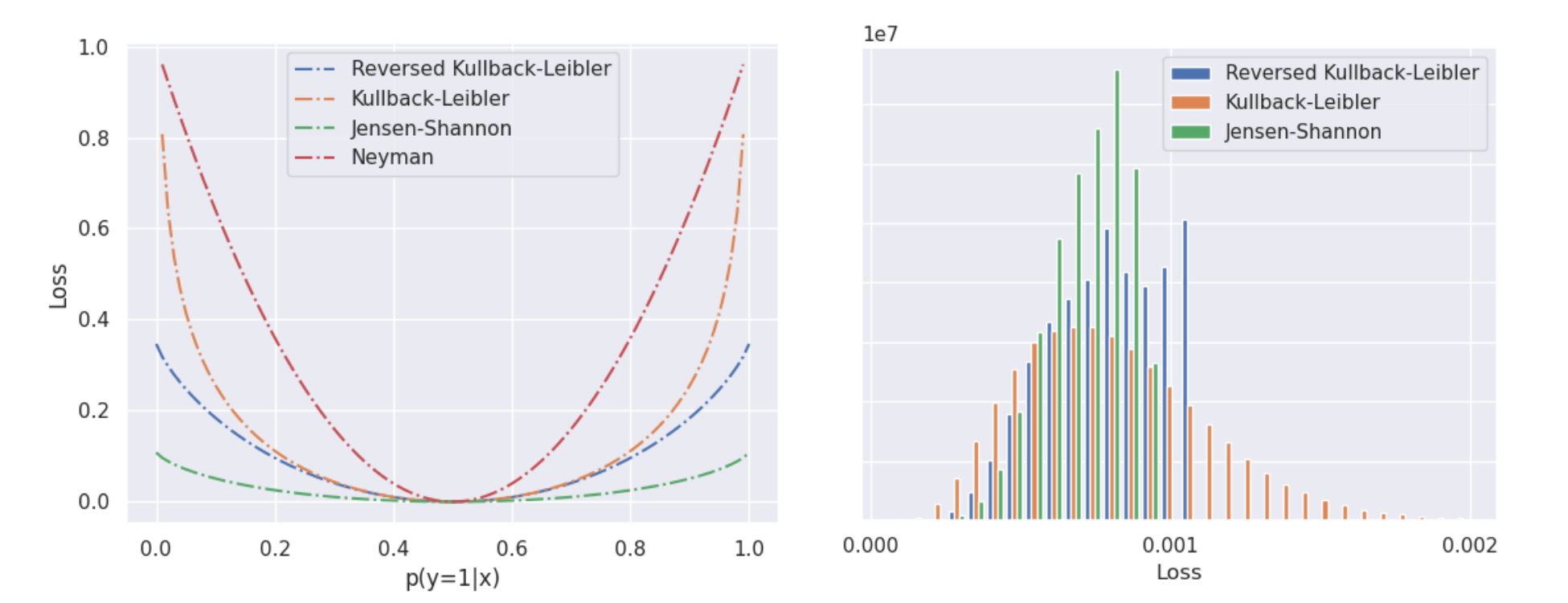}
    \caption{Left: f-divergences towards the uniform distribution in a two-class setup.
Jensen-Shannon offers the most robust response.
Right: Histograms of $\lambda L_{\mathrm{neg}}$ in synthetic negatives 
at the beginning of joint fine-tuning.
The modulation factors $\lambda$
have been separately validated
for each of the three choices of $L_{\mathrm{neg}}$.
Jensen-Shannon divergence produces a more uniform learning signal than other f-divergences and avoids a high variety of $L_{\mathrm{neg}}$.}
    \label{fig:theory-ood-1}
\end{figure}

The left part of Figure \ref{fig:theory-ood-1} plots several f-divergences in the two-class setup.
We observe that the Jensen-Shannon divergence mildly penalizes high-confidence predictions, which makes it a suitable candidate for a robust loss.
Such behaviour promotes graceful performance degradation in cases of errors of the generative model.
The right part of Figure \ref{fig:theory-ood-1} visualizes a histogram of per-pixel loss while fine-tuning our model on road-driving images.
The figure shows that the histogram of JS divergence has fewer high-loss pixels than the other f-divergence candidates.
Long tails of the KL divergences (forward and reverse) indicate a very high loss in pixels that resemble inliers.
As hinted before, these pixels give rise to very high gradients with respect to the parameters of the discriminative model.
These gradients may override the impact of the standard discriminative loss $L_\text{disc}$, and lead to high-entropy discriminative predictions that disrupt our anomaly score and lead to false positive predictions.  
Consequently, we formulate $L_\text{neg}$ in terms of JS divergence between the uniform distribution over classes and the softmax output:
\begin{equation}
\label{eq:lneg}
    L^{ij}_\text{neg}  = \text{JS}(\text{U}, p_\theta(\mathbf{y}^{ij}|\mathbf{x}'))
\end{equation}

\subsection{Outlier-aware inference with divergence-based scoring}
\label{sec:div-inf}

\begin{figure}[ht]
    \centering
    \includegraphics[width=\linewidth]{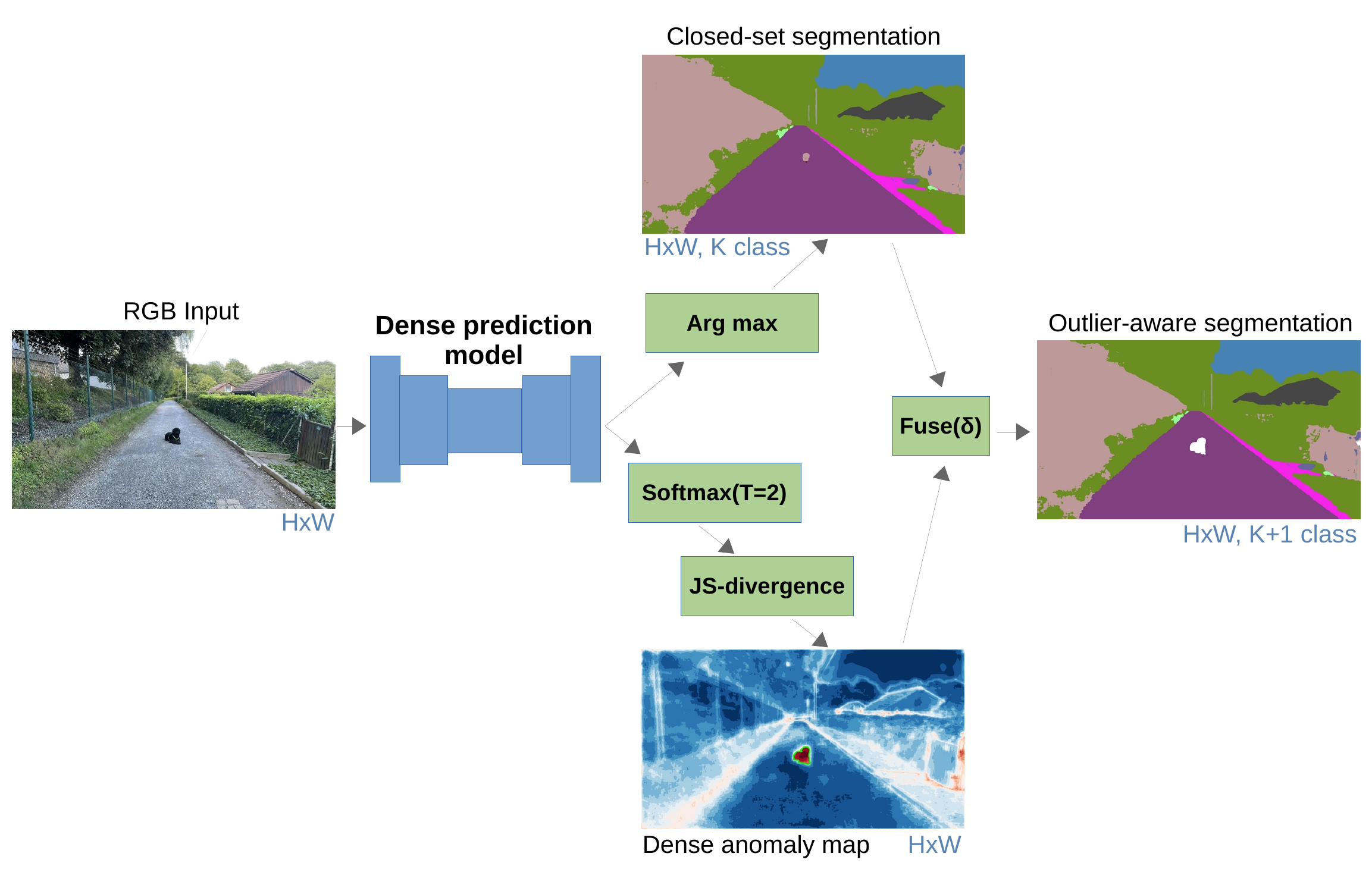}
    \caption{Dense outlier-aware inference. 
We infer dense logits with a closed-set model.
We recover the dense OOD map according to our divergence-based score (JSD).
Closed-set predictions are overridden in the outlier-aware output wherever the OOD score exceeds the threshold $\delta$.}
    \label{fig:infrence-scheme}
\end{figure}

Figure \ref{fig:infrence-scheme} summarizes inference according to the proposed method for outlier-aware semantic segmentation.
The input image is fed into the discriminative model. 
The produced logits are fed into two branches.
The top branch delivers closed-set predictions through arg-max.
The bottom branch recovers the dense OOD map through temperature scaling, softmax and JS divergence with respect
to the uniform distribution.
Our dense OOD score at every pixel $i,j$ reflects the $L_{\mathrm{neg}}$ loss (\ref{eq:lneg}):
\begin{equation}
    s^{ij}(\mathbf{x}) = \text{JS}(\text{U}, \text{softmax}(\mathbf{l}^{ij}/T)).
\end{equation}
U stands for uniform distribution over inlier classes, $\mathbf{l}$ represents logits while $T$ is a temperature hyperparameter.
The two branches are fused into the final outlier-aware segmentation map.
The OOD map overrides the closed-set prediction wherever the OOD score exceeds a dataset-wide threshold.

Temperature scaling \cite{guo17icml} reduces the relative OOD score of distributions with two dominant logits as opposed to distributions with homogeneous non-maximum logits.
This discourages false positive OOD responses at semantic borders.
We use the same temperature T=2 in all experimental comparisons with respect to previous methods.   
Note that our inference is very fast since we use our generative model  only to simulate anomalies during training. 
This is different from image resynthesis \cite{lis19iccv} and embedding density \cite{blum21ijcv} where the generative model has to be used during inference.
Next, we compare the distributional coverage of synthetic negatives generated by normalizing flow with respect to their GAN-generated counterparts.

\subsection{
Coverage-oriented generation of synthetic negatives 
}
\label{sec:theoretical_analysis} 
We provide a theoretical argument
that our synthetic negatives provide
a better distribution coverage
than their GAN counterparts \cite{lee18iclr}.
Our argument proceeds by analyzing 
the gradient of the joint loss 
with respect to the generator of synthetic negatives for both approaches.
For brevity, we omit the spatial locations and loss modulation hyperparameters.

Adversarial outlier-aware learning \cite{lee18iclr} jointly optimizes 
the zero-sum game between the generator $G_\psi$ 
and the discriminator $D_\phi$, closed-set classification $P_\theta$, and the confidence objective that enforces uncertain classification 
in the negative data points \cite{lee18iclr}:
\begin{multline}
\label{eq:gan_joint}
    L_{\mathrm{adv}}(\phi, \psi, \theta) =\\ \int p^*(\mathbf{x}) \ln D_\phi(\mathbf{x}) \, d\mathbf{x} + \int p_{G_\psi}(\mathbf{x}) \ln (1- D_\phi(\mathbf{x})) \, d\mathbf{x} \\
    - \int p^*(y,\mathbf{x}) \ln P_\theta(y|\mathbf{x}) \, dy\,d\mathbf{x} + \int p_{G_\psi}(\mathbf{x}) \mathcal{F}(P_\theta, \text{U}) \, d\mathbf{x}.
\end{multline}
We denote the true data distribution as $p^*$ while
$\mathcal{F}$ corresponds to the chosen f-divergence.
The gradient of the joint loss (\ref{eq:gan_joint}) w.r.t.\ the generator parameters $\psi$ vanishes in the first and the third term.
The remaining terms enforce that the generated samples fool the discriminator and yield high-entropy closed-set predictions:
\begin{multline}
\label{eq:grad_gan_joint}
   \frac{\partial L_{\mathrm{adv}}(\phi, \psi, \theta)}{\partial \psi} = \frac{\partial}{\partial \psi} \int p_\psi(\mathbf{x}) \ln (1- D_\phi(\mathbf{x})) \, d\mathbf{x} \\
    + \frac{\partial}{\partial \psi} \int p_\psi(\mathbf{x}) \mathcal{F}(P_\theta, \text{U}) \, d\mathbf{x}.
\end{multline}
However, fooling the discriminator does not imply distributional coverage.
In fact, the adversarial objective may cause mode collapse \cite{metz17iclr} which is detrimental to sample  variability.

Our joint learning objective (\ref{eq:joi_e2e}) optimizes the likelihood of inlier samples, the closed-set classification loss, and low confidence in synthetic negatives:
\begin{multline}
\label{eq:mle_joint}
    L(\gamma, \theta) = -\int p^*(\mathbf{x}) \ln p_\gamma(\mathbf{x}) \, d\mathbf{x} \\
    - \int p^*(y,\mathbf{x}) \ln P_\theta(y|\mathbf{x}) \, dy\,d\mathbf{x} + \int p_\psi(\mathbf{x}) \mathcal{F}(P_\theta, \text{U}) \, d\mathbf{x}.
\end{multline}
The gradient of the loss (\ref{eq:mle_joint}) w.r.t.\ the normalizing flow parameters $\gamma$ vanishes in the second term.
The remaining terms enforce that the generated samples cover all modes of $p^*$ and, as before, yield high-entropy discriminative predictions:
\begin{multline}
\label{eq:grad_mle_joint}
   \frac{\partial L(\gamma, \theta)}{\partial \gamma} = - \frac{\partial}{\partial \gamma} \int p^*(\mathbf{x}) \ln p_\gamma(\mathbf{x}) \, d\mathbf{x} \\
    + \frac{\partial}{\partial \gamma} \int p_\gamma(\mathbf{x}) \mathcal{F}(P_\theta, \text{U}) \, d\mathbf{x}.
\end{multline}
The resulting gradient entices the generative model to produce samples along the border of the inlier distribution.
Consequently, we say that our synthetic negatives are coverage-oriented.
The presented analysis holds for any generative model that optimizes
the density of the training data.
Experimental evaluations in the following sections provide conclusive empirical confirmation 
for the advantages of synthetic negatives generated by normalizing flow (cf.\ Table \ref{tbl:abl-gm}).

\section{Experimental setup}

This section describes our experimental setup for dense out-of-distribution detection.
We review the employed datasets, introduce performance metrics, and
describe the training details.

\subsection{Benchmarks and Datasets}

\begin{figure}[ht]
    \centering
    \includegraphics[width=\linewidth]{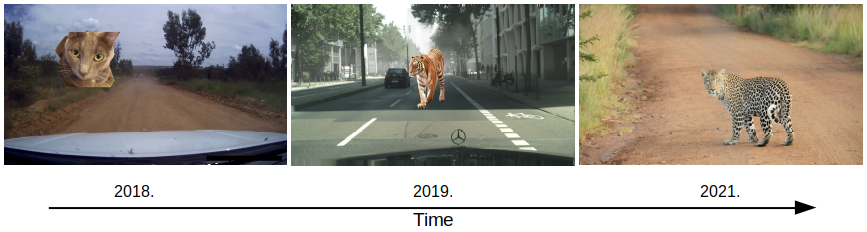}
    \caption{Development of dense OOD detection in road-driving} through time.
    Early work pastes objects at random locations \cite{bevandic18arxiv}.
    This was  improved by carefully choosing pasting locations and postprocessing \cite{blum21ijcv}. 
    Recent work ensures outliers match the environment by selecting real-world scenes \cite{chan21arxiv}.
    \label{fig:eval_progress}
\end{figure}

Benchmarks for dense OOD detection in road-driving scenes have experienced substantial progress in recent years (cf.\ Figure \ref{fig:eval_progress}).
In parallel, significant effort has been invested into artificial datasets by leveraging simulated environments \cite{hendrycks19arxiv,maag22accv}.
Similarly, remote-sensing segmentation datasets have grown  both in size in complexity \cite{carvalho22rs}.

\textbf{WD-Pascal}
\cite{bevandic18arxiv} has been created by pasting Pascal VOC \cite{everingham10ijcv} objects into WildDash \cite{zendel18eccv} images.
The resulting dataset allows for evaluating outlier detection in demanding conditions.
However, the random pasting policy disturbs the scene layout as shown in Figure \ref{fig:eval_progress} (left).
Consequently, there is a concern that such anomalies may be easier to detect.

\textbf{Fishyscapes} \cite{blum21ijcv} evaluates model's ability to detect outliers in urban driving scenarios.
The benchmark consists of two datasets: FS LostAndFound and FS Static.
FS LostAndFound is a small subset of original LostAndFound \cite{pinggera16iros} which contains small objects on the roadway (e.g.\ toys, boxes or car parts that could fall off).
FS Static contains Cityscapes validation images overlaid with Pascal VOC objects.
The objects are positioned according to the perspective and further postprocessed to obtain smoother OOD injection.

\textbf{SegmentMeIfYouCan} (SMIYC) \cite{chan21arxiv} quantifies dense outlier detection performance in multiple setups.
The benchmark consists of three datasets: AnomalyTrack, ObstacleTrack and LostAndFound \cite{pinggera16iros}.
AnomalyTrack provides large anomalous objects which are fully aligned with the environment.
For instance, they have a leopard in the middle of a dirt road as shown in Fig.\ \ref{fig:eval_progress} (right).
LostAndFound \cite{pinggera16iros} tests detection of small hazardous objects (e.g.\ boxes, toys, car parts, etc.) in urban scenes.
Finally, ObstacleTrack tests detection of small objects on various road types.
Inconsistent road surfaces can trick the detector and increase the false positive rate.
ObstacleTrack and LostAndFound measure OOD detection performance solely on the driving surface while AnomalyTrack considers the detection across the whole image.
Consequently, SMIYC provides a solid notion on OOD segmentation performance of a model deployed in the wild.

\textbf{StreetHazards} \cite{hendrycks19arxiv} is a synthetic dataset created with the CARLA game engine.
The dataset captures simulated urban environment with carefully inserted anomalous objects (e.g.\ a horse carriage or a helicopter on the road).
Simulating anomalies in virtual environments is appealing due to high flexibility in positioning and appearance, as well as low cost of data accumulation.
Unfortunately, there is a notable quality mismatch between simulated environments and the real world.
Still, this approach has great potential for evaluating outlier-aware segmentation due to cheap ground truth with K+1 classes.
We use StreetHazards for measuring outlier-aware segmentation performance according to open-mIoU \cite{grcic22eccv}.

\textbf{BSB} \cite{carvalho22rs} is a remote sensing dataset with aerial images of Brasilia. 
It contains 3400 labeled images of 512$\times$512 pixels. 
The official split designates 3000 train, 200 validation and 200 test images.
The labels include 3 stuff classes (street, permeable area, and lake) 
and 11 thing classes (e.g.\ swimming pool, vehicle, sports court).
We extract \textit{boat} and \textit{harbour} into the OOD test set.
The resulting BSB-OOD dataset contains 2840 training images with 12 inlier classes, while the OOD test set contains 184 images.
This setup is similar to \cite{hendrycks19arxiv,grcic21visapp,vaze22icmlshw} that also select a subset of classes as OOD samples.
Note that there are other remote sensing datasets such as Vaihingen and Potsdam from the International Society for Photogrammetry and Remote Sensing (ISPRS).
However, these datasets have fewer labels and an order of magnitude fewer images.
Also, the So2Sat LCZ42 dataset \cite{zhu19arxiv} contains only small-resolution images and image-level labels.
Hence, we opt for larger dataset and better performance estimates.

\subsection{Metrics}
\label{sec:metrics}

We measure OOD segmentation performance using average precision (AP) \cite{everingham10ijcv}, false-positive rate at true-positive rate of 95\% ($\mathrm{FPR}_{95}$) \cite{hendrycks17iclr} and AUROC.
AP is well suited for measuring OOD detection performance since it emphasizes the minority class \cite{davis06icml,blum21ijcv,chan21arxiv}.
A perfect OOD detector would have AP equal to one.
Likewise, $\mathrm{FPR}_{95}$ is significant for real-world applications since high false-positive rates would require a large number of human interventions in practical deployments and therefore severely diminish the practical value of an autonomous system.
We measure outlier-aware segmentation performance by open-mIoU \cite{grcic22eccv}.
Open-mIoU penalizes outliers being recognized as inliers and inliers being wrongly detected as outliers.
Compared to mIoU over K+1 classes, open-mIoU does not count true positive outlier predictions and averages over K instead of K+1 classes.
Open-mIoU performance of an outlier-aware segmentation model with ideal OOD detection would be equal to the closed-set mIoU of the same model.
Hence, the difference between the two metrics quantifies the performance gap caused by the presence of outliers \cite{grcic22eccv}.

\subsection{Implementation details}
\label{subsec:experimental-setup}
All our models are based on Ladder DenseNet-121 (LDN-121) 
due to memory efficiency and fast experimentation \cite{kreso21tits}.
However, our framework can accommodate any other dense prediction architecture.
All our experiments consist of two training stages.
In both stages we utilize Cityscapes \cite{cordts16cvpr}, Vistas \cite{neuhold17iccv} and Wilddash 2 \cite{zendel18eccv}.
These three datasets contain $25\,231$ images.
The images are resized to 1024 pixels (shorter edge), randomly flipped with the probability of 0.5, randomly resized in the interval $[0.5, 2]$, and randomly cropped to $768 \times 768$ pixels.
We optimize our models with Adam.
In the first stage we train for 25 epochs without synthetic negatives.
We use batch size 16 as validated in previous work \cite{kreso21tits}.
The starting learning rate is set to $10^{-4}$ for the feature extractor and $4\cdot 10^{-4}$ for the upsampling path.
The learning rate is annealed according to a cosine schedule to the minimal value of $10^{-7}$ which would have been reached in the 50th epoch.

In the second stage, we train for 15 epochs on mixed-content images (cf.\ section \ref{sec:neg_ds_to_flow}).
In this stage, we use a batch size of 12 due to limited GPU memory.

We did not use gradient accumulation due to batch normalization layers.
Instead, we opted for gradient checkpointing \cite{bulo18cvpr,barron19cvpr,kreso21tits}.
The initial learning rate is set to $1\cdot 10^{-5}$ for the upsampling path and $2.5\cdot 10^{-6}$ for the backbone.
Once more the learning rate is decayed according to the cosine schedule to the value of $10^{-7}$.
We set the hyperparameter $\lambda$ to $3 \cdot 10^{-2}$.
This value is chosen so that the closed-set segmentation performance is not reduced.

We generate rectangular synthetic samples
with dimensions from $\mathcal{U}(16,216)$
by leveraging DenseFlow-25-6 \cite{grcic21neurips}.
The flow is pretrained on random $64 \times 64$ crops from Vistas.
We train the flow with the Adamax optimizer with learning rate set to $10^{-6}$.
In the case of WD-Pascal, we train our model only on Vistas in order to achieve a fair comparison with the previous work \cite{bevandic19gcpr}.
In the case of StreetHazards, we train on the corresponding train subset for 80 epochs on inlier images and 40 epochs on mixed-content images.
In the case of Fishyscapes, we train exclusively on Cityscapes.
We train for 150 epochs during stage 1 (inliers) and 50 epochs during
stage 2 (mixed content).
In the case of BSB-OOD dataset, we train LDN-121 for 150 epochs with a batch size of 16 on inlier images and then fine-tune on mixed-content images for 40 epochs.
We sample synthetic negatives with dimensions from  $\mathcal{U}(16,64)$.
The flow was pre-trained on $32 \times 32$ random inlier crops of BSB-OOD images for 2k epochs with batch size of 256.
All other hyperparameters are kept constant across all experiments.
Each experiment lasts for approximately 38 hours on a single NVIDIA RTX A5000.

\section{Experimental evaluation}

We evaluate OOD detection performance of NFlowJS on 
road-driving scenes and aerial images.
Experiments on road-driving images suggest that our synthetic negatives can deliver comparable performance to real negatives (Sec.\ \ref{sec:exp_ood_rd}).
Furthermore, our synthetic negatives become a method of choice in setups 
with a large domain gap towards candidate datasets for sourcing real negative training samples (Sec.\ \ref{sec:exp_ood_rs}).

We compare our performance with respect to contemporary methods which do not require the negative dataset or image resynthesis.
Still, we list all methods in our tables, so we can discuss our method in a broader context.
We also analyze the sensitivity of our method with respect to the distance of the OOD object from the camera.
Finally, we measure the computational overhead of our method with respect to the baseline and visualize our synthetic samples.

\subsection{Dense out-of-distribution detection in road-driving scenes}
\label{sec:exp_ood_rd}

Table \ref{tbl:wd-pascal} presents performance on WD-Pascal averaged over 50 runs \cite{bevandic19gcpr}.
All methods have been trained on Vistas datasets and achieve similar mIoU performance.
Column Aux data indicates whether 
the method trains on real negative data.
We choose ADE20k for this purpose since it offers instance-level ground truth.
The bottom section compares our method
with early approaches:
MC dropout \cite{kendall17nips},
ODIN \cite{liang18iclr},
and max-softmax \cite{hendrycks17iclr}.
These approaches are not competitive with the current state-of-the-art.
The top section shows that training with auxiliary negative data
can significantly improve performance.
However, our method closes the performance gap.
It outperforms all other methods in FPR95 and AUROC metrics
while achieving competitive AP.
\begin{table}[ht]
\centering
\caption{
Performance evaluation on WD-Pascal \cite{bevandic19gcpr}.
}
\label{tbl:wd-pascal}
\begin{footnotesize}
\begin{tabular}{l|cccc}
\hline \hline
Method & Aux data & AP  $\uparrow$          & $\mathrm{FPR}_{95}$   $\downarrow $     & AUROC  $\uparrow $      \\ \hline
OOD head \cite{bevandic19gcpr} &  \cmark  &  \textbf{34.9} $\pm$ 6.8 & 40.9 $\pm$ 3.9  & 88.8  $\pm$ 1.6  \\
MSP \cite{bevandic19gcpr} &  \cmark  &  33.8 $\pm$ 5.1  & 35.5 $\pm$ 3.4 &  91.1 $\pm$ 1.0  \\
Void Cls. \cite{blum21ijcv} &  \cmark  &  25.6 $\pm$ 5.5  &  44.2 $\pm$ 4.7 &  87.7 $\pm$ 1.7  \\\hdashline
MC Drop. \cite{kendall17nips} & \xmark   &  9.7 $\pm$ 1.2  & 41.1 $\pm$ 3.7  &   86.0 $\pm$ 1.2\\
ODIN \cite{liang18iclr} & \xmark   & 6.0 $\pm$ 0.5   & 53.7 $\pm$ 7.0  &  79.9 $\pm$ 1.5  \\
MSP \cite{hendrycks17iclr} &  \xmark  & 5.0 $\pm$ 0.5  & 48.8 $\pm$ 4.7  & 78.7 $\pm$ 1.5   \\
NFlowJS (ours) & \xmark   &  \textbf{30.2} $\pm$ 4.1  &  \textbf{32.3} $\pm$ 5.9 &  \textbf{92.3} $\pm$ 1.3 \\\hline
\end{tabular}
\end{footnotesize}
\end{table}

\begin{table*}[ht]
\centering
 \caption{Dense out-of-distribution detection performance on SegmentMeIfYouCan and Fishyscapes.
}
\label{tbl:smiyc}
\begin{footnotesize}
\begin{tabular}{l|ccccccccccccc}
\hline\hline
\multirow{3}{*}{Method} & & & \multicolumn{6}{|c|}{SegmentMeIfYouCan \cite{chan21arxiv}} &  \multicolumn{5}{c}{Fishyscapes \cite{blum21ijcv}} \\\cline{4-14}
 & \multicolumn{1}{c}{\multirow{2}{*}{Aux}} & \multicolumn{1}{c}{\multirow{2}{*}{Img}} & \multicolumn{2}{|c|}{AnomalyTrack} & \multicolumn{2}{c|}{ObstacleTrack}
& \multicolumn{2}{c|}{LAF-noKnown} & \multicolumn{2}{c|}{FS LAF} & \multicolumn{2}{c|}{FS Static} & CS val\\
 & \multicolumn{1}{c}{data} & \multicolumn{1}{c}{rsyn.} & \multicolumn{1}{|c}{AP} & \multicolumn{1}{c|}{$\mathrm{FPR}_{95}$} & AP & \multicolumn{1}{c|}{$\mathrm{FPR}_{95}$} & AP & \multicolumn{1}{c|}{$\mathrm{FPR}_{95}$}  & AP & \multicolumn{1}{c|}{$\mathrm{FPR}_{95}$} & AP & \multicolumn{1}{c|}{$\mathrm{FPR}_{95}$} & $\overline{\mathrm{IoU}}$ \\ \hline \hline
SynBoost \cite{biase21cvpr} & \cmark& \cmark & 56.4 & 61.9 & 71.3 & 3.2 & 81.7 & 4.6 & \textbf{43.2} & 15.8 & 72.6 & 18.8 & 81.4 \\
Prior Entropy \cite{malinin18nips} & \cmark & \xmark & - & - & - & - & - & - & 34.3 & 47.4 & 31.3 & 84.6 & 70.5 \\
OOD Head \cite{bevandic19gcpr} & \cmark & \xmark & - & - & - & - & - & -  & 31.3 & 19.0 & \textbf{96.8} & \textbf{0.3} & 79.6 \\
Void Classifier \cite{blum21ijcv} & \cmark& \xmark & 36.6 & 63.5 & 10.4 & 41.5 & 4.8 & 47.0 & 10.3 & 22.1 & 45.0 & 19.4 & 70.4 \\
Image Resyn. \cite{lis19iccv} & \xmark& \cmark& 52.3 & 25.9 & 37.7 & 4.7 & 57.1 & 8.8 & 5.7 & 48.1 & 29.6 & 27.1 & 81.4  \\
Road Inpaint. \cite{lis20arxiv} & \xmark& \cmark & - & -  & 54.1 & 47.1 & 82.9 & 35.8  & - & - & - & - & -  \\\hdashline
Max softmax \cite{hendrycks17iclr} & \xmark&  \xmark & 28.0 & 72.1 & 15.7 & 16.6  &  30.1 & 33.2 &  1.8 & 44.9 & 12.9 & 39.8 & 80.3 \\
MC Dropout \cite{kendall17nips} & \xmark& \xmark & 28.9 & 69.5 & 4.9 & 50.3 & 36.8 & 35.6 & - & - & - & - & - \\
ODIN \cite{liang18iclr} & \xmark&  \xmark & 33.1 & 71.7 & 22.1 & 15.3 & 52.9 & 30.0 & - & - & - & - & -  \\
SML \cite{jung21iccv} & \xmark & \xmark & - & - & - & - & - & -   & -  & 31.7 & 21.9 & 52.1 & 20.5 \\
Embed.\ Dens.\ \cite{blum21ijcv} & \xmark&  \xmark & 37.5 & 70.8 & 0.8 & 46.4 & 61.7 & 10.4 & 4.3 & 47.2 & \textbf{62.1} & 17.4 & 80.3\\
JSRNet \cite{vojir21iccv} & \xmark&  \xmark & 33.6& 43.9 & 28.1 & 28.9 & 74.2 & 6.6 & - & - & - & - & - \\
NFlowJS (ours) & \xmark&  \xmark & \textbf{56.9} & \textbf{34.7} & \textbf{85.5} & \textbf{0.4} & \textbf{89.3} & \textbf{0.7} &\textbf{39.4} & \textbf{9.0} & 52.1 & \textbf{15.4} & 77.4 \\ \hline
\end{tabular}
\end{footnotesize}
\end{table*}

Table \ref{tbl:smiyc} presents performance evaluation on SMIYC \cite{chan21arxiv} and Fishyscapes \cite{blum21ijcv}.
Our method outperforms all previous methods on AnomalyTrack, ObstacleTrack as well as LAF-noKnown.
We achieve such results despite refraining from image resynthesis \cite{lis19iccv,lis20arxiv,biase21cvpr}, partial image reconstruction \cite{vojir21iccv} or training on real negative images \cite{blum21ijcv}.
Our method achieves very low  $\mathrm{FPR}_{95}$ (less than 1\%) on ObstacleTrack and LostAndFound-noKnown.
This is especially important for real-world applications where high incidence of false positives may make OOD detection useless.
Note that ObstacleTrack includes small obstacles in front of a variety of road surfaces, 
which makes it extremely hard not to misclassify road parts as anomalies.
Moreover, this dataset includes low-visibility images captured at dusk and other challenging evaluation setups.
Our synthetic negative data also achieve competitive performance on FS LostAndFound. 
Our method outperforms others in terms of $\mathrm{FPR}_{95}$ while achieving the second best AP.
We slightly underperform only with respect to SynBoost which trains on real negative data and precludes real-time inference due to image resynthesis.
In the case of FS Static dataset, our method achieves the best $\mathrm{FPR}_{95}$ and the second best AP among the methods which do not train on auxiliary data.

We have also applied our method to a pre-trained third-party closed-set model and submitted the results to the Fishyscapes benchmark.
We have chosen a popular DeepLabV3+ model which achieves high performance due to training on unlabeled video data \cite{zhu19cvpr}.
This choice promotes fair comparison, since the same model has also been used in several other benchmark submissions \cite{jung21iccv,vojir21iccv}.
Please note that we use parameters which have not been trained on Cityscapes val in order to allow fair evaluation on FS Static.
The corresponding dense OOD detection model achieves 
43.7 AP and 8.6 $\mathrm{FPR}_{95}$ on FS LAF,  
54.7 AP and 10.0 $\mathrm{FPR}_{95}$ on FS Static, while having 
80.7 mIoU on Cityscapes val. 
We do not show these results in Table \ref{tbl:smiyc} in order to keep the same model across all assays.
This result clearly shows that our method can also be applied to third-party models and deliver strong results.

\begin{figure}[ht]
    \centering
    \includegraphics[width=\linewidth]{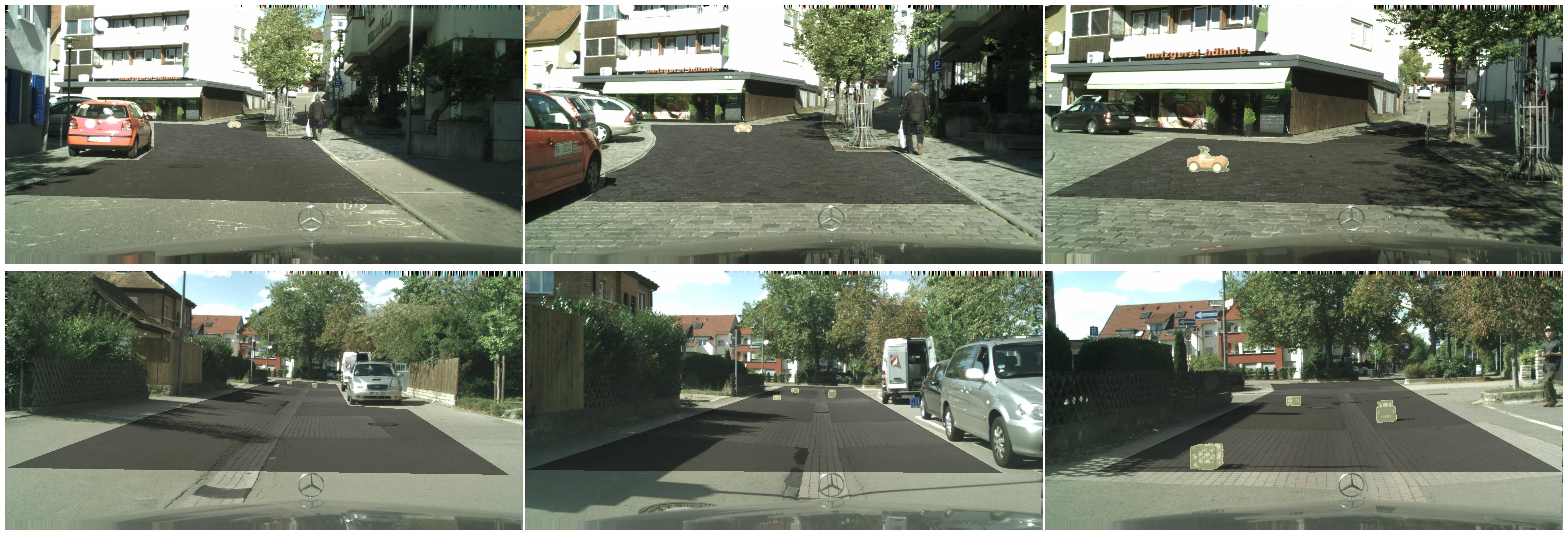}
    \caption{OOD detection on LostAndFound dataset. Our method can detect obstacles at different distances from the camera (top) as well as multiple obstacles in one image (bottom). Road ground truth is designated in grey and the predicted OOD in yellow. Please zoom in to see the distant obstacles}.
    \label{fig:lf-ood}
\end{figure}

Figure \ref{fig:lf-ood} shows qualitative performance on two sequences of images from SMIYC LostAndFound.
Road ground truth is designated in grey and the detected obstacles are in yellow.
The top sequence contains obstacles which change position through time.
The bottom sequence contains multiple anomalous objects.
Our method succeeds to detect a toy car and cardboard boxes even though no such objects were present during the training.
Column 1 contains distant obstacles so please zoom in for better visibility.

Table \ref{table:osr_sh}  shows OOD detection and outlier-aware semantic segmentation on StreetHazards.
We produce outlier-aware semantic predictions  by correcting closed-set predictions with our dense OOD map (Sec.\ \ref{sec:div-inf}).
We validate the OOD threshold in order to achieve TPR=95\%  \cite{grcic22eccv} and measure performance according to mIoU over K+1 classes as well as with open-mIoU \cite{grcic22eccv}.
To the best of our knowledge, our method outperforms all previous work.
In particular, our method is better than  methods which utilize auxiliary negative datasets \cite{hendrycks19iclr,liu20neurips,bevandic19gcpr} and the method based on image resynthesis \cite{xia20eccv}.
We note that there is still a significant performance degradation in presence of outliers.
Closed-set performance is more than 65\% mIoU, while outlier-aware performance peaks at 45\%.
Future research should strive to close this gap to provide safer segmentation in the wild. 
\begin{table}[ht]
\centering
\caption{Performance evaluation on StreetHazards \cite{hendrycks19arxiv}.
}
\label{table:osr_sh}
\begin{tabular}{lcccccc}
\hline \hline
\multirow{2}{*}{Method} & \multicolumn{1}{c|}{Aux.} & \multicolumn{2}{c|}{Anomaly det.} & \multicolumn{1}{c|}{Closed} & \multicolumn{2}{c}{Open}\\
  & \multicolumn{1}{c|}{data}  &  AP         & \multicolumn{1}{c|}{$\mathrm{FPR}_{95}$}     & \multicolumn{1}{c|}{$\overline{\mathrm{IoU}}$} &  $\overline{\mathrm{IoU}}$ & o-$\overline{\mathrm{IoU}}$ \\ \hline \hline
SynthCP \cite{xia20eccv} & \xmark &  9.3           & 28.4      & - &  -  & -\\
TRADI \cite{franchi20eccv} & \xmark &  7.2           & 25.3       & - & -  & - \\
OVNNI \cite{franchi20arxiv} & \xmark & 12.6  & 22.2  & 54.6 & - & - \\
SO+H \cite{grcic21visapp}& \xmark & 12.7  & 25.2  & 59.7 & - & - \\
DML \cite{cen21iccv} & \xmark  & 14.7  & 17.3  &  - & - & - \\
MSP \cite{hendrycks17iclr} & \xmark  &  7.5   &   27.9 & 65.0 & 32.4 & 35.1\\
Max logit \cite{hendrycks19arxiv} & \xmark  & 11.6   &  22.5   & 65.0  & 38.0 & 41.2\\
ODIN \cite{liang18iclr}&   \xmark   &    7.0       & 28.7 & 65.0  & - & 28.8\\
ReAct \cite{sun21neurips} & \xmark & 10.9  & 21.2 & 62.7 & 31.8 &  34.0\\
Energy \cite{liu20neurips}& \cmark &  12.9  & 18.2  & 63.3 & 39.6 &  42.7\\
MSP + OE \cite{hendrycks19iclr} & \cmark  &  14.6   &  17.7   & 61.7 & 40.8  & 43.8\\
OOD-Head \cite{bevandic19gcpr}  & \cmark   &  19.7 &  56.2   &  66.6 & - & 33.9\\
OH*MSP \cite{bevandic22ivc} & \cmark  & 18.8  & 30.9  &   \textbf{66.6}  & -  & 43.6\\
NFlowJS (ours) & \xmark &  \textbf{22.2}  & \textbf{16.2}  &  65.0 & \textbf{41.6} & \textbf{44.9} \\\hline
\end{tabular}
\end{table}

We implemented \cite{hendrycks19iclr,liang18iclr,liu20neurips,sun21neurips} into our codebase according to official implementations.
For the energy fine-tuning, we have conducted hyperparameter search as suggested in \cite{liu20neurips}: $m_{in} \in \{-15, -23, -27\}$ and $m_{out} \in \{-5, -7\}$.
The optimal values for dense setup are $m_{in}=-15$ and $m_{out}=-5$.
We have validated ReAct \cite{sun21neurips} for $c \in \{0.9, 0.95, 0.99\}$.
The best results are obtained with $c=0.99$.

\begin{figure}[ht]
    \centering
    \includegraphics[width=0.95\linewidth]{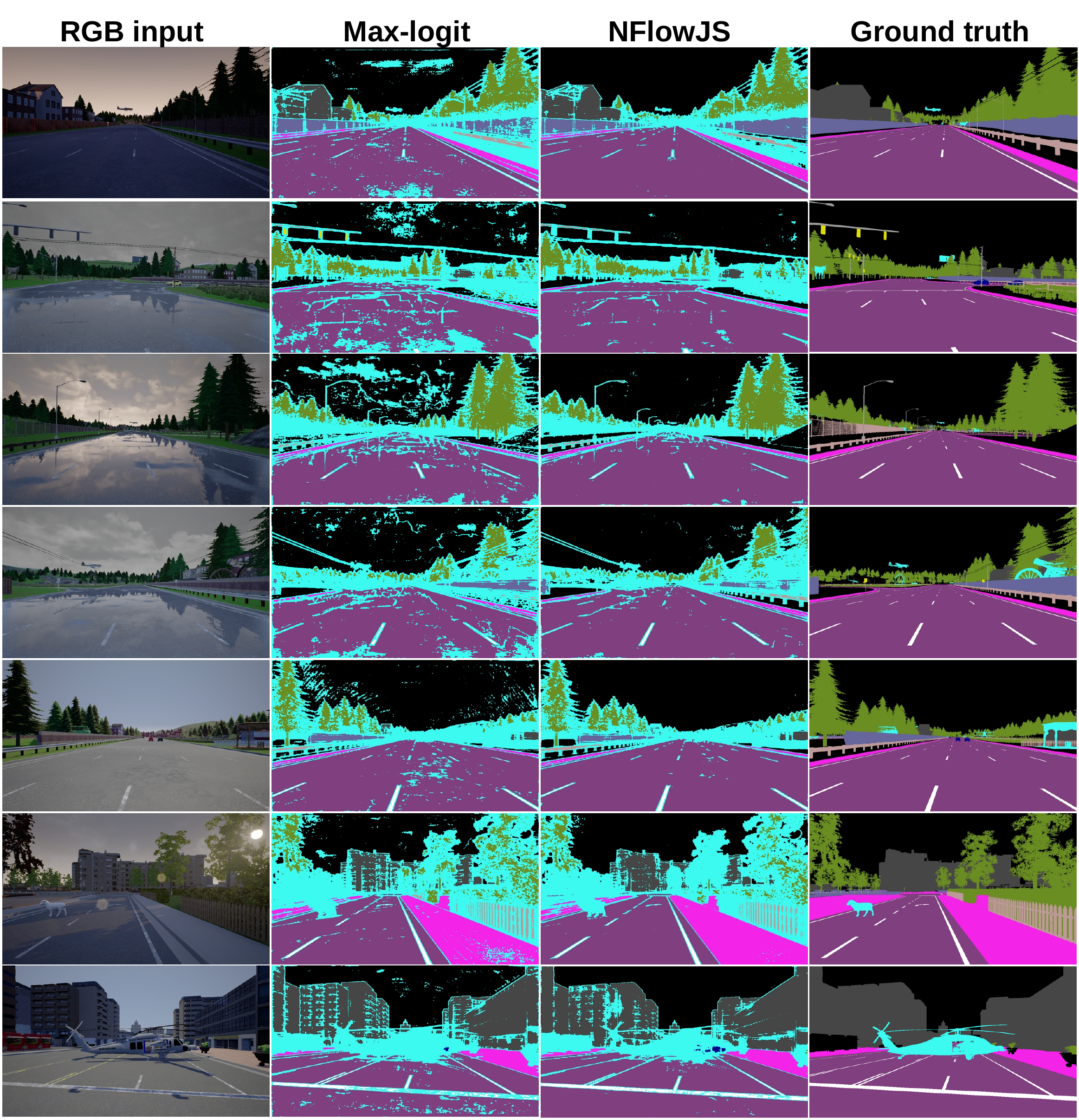}
    \caption{Outlier-aware segmentation on StreetHazards.
    The detected outliers are marked with cyan.
    Our method reduces the number of false positives over max-logit baseline. Still, further research is required in order to achieve closed-set performance in presence of outliers.}
    \label{fig:sh_osr}
\end{figure}

Figure \ref{fig:sh_osr} compares outlier-aware semantic segmentation performance of the proposed method with respect to the max-logit baseline \cite{hendrycks19arxiv} on StreetHazards.
Anomalous pixels are designated in cyan.
Our method reduces the number of false positives.
However, safe and accurate outlier-aware segmentation is still an open problem.

\subsection{Dense out-of-distribution detection in remote sensing}
\label{sec:exp_ood_rs}

We compare our method with standard baselines for OOD detection \cite{hendrycks17iclr,liu20neurips,hendrycks19iclr} as well as with methods specifically developed for OOD detection in remote sensing imagery \cite{gawlikowski21igrass,dasilva20arxiv}.
Table \ref{tbl:bsb_ood} shows the performance on the BSB-aerial-OOD dataset \cite{carvalho22rs}.
Some methods train on real negative data (cf.\ Aux data).
The top section presents several OOD detection baselines.
We observe that training with real negative samples outperforms the MSP baseline \cite{hendrycks17iclr} but underperforms with respect to our synthetic samples.
This is not surprising since the pasted negative instances involve a  different camera perspective than aerial imagery.
The middle section presents methods that are explicitly designed for aerial images.
Morph-OpenPixel (MOP) \cite{dasilva20arxiv} erodes the prediction confidence at object boundaries with morphological filtering.
Morphological filtering improves $\text{FPR}_{95}$ but impairs AP with respect to the MSP baseline.
$\text{DPN}^-$ \cite{gawlikowski21igrass} achieves runner-up AUROC and $\text{FPR}_{95}$ performance.
The bottom part shows the performance of our model.
JSDiv is the same as NFlowJS except that it
uses negatives from ADE20k instead of synthetic ones.
NFlowJS generates dataset-specific negatives along the border between the known and the unknown.
NFlowJS outperforms methods which train on real negative data, indicating that synthetic negatives may be a method of choice when an appropriate negative dataset is unavailable.

\begin{table}[ht]
\centering
\caption{
Performance evaluation on  images from BSB-OOD.
}

\label{tbl:bsb_ood}
\begin{footnotesize}
\begin{tabular}{l|cccc}
\hline \hline
Method & Aux & AP  & $\mathrm{FPR}_{95}$  & AUROC    \\ \hline
MSP \cite{hendrycks17iclr} &  \xmark  &  35.1 & 13.5 &  96.4  \\
MSP + OE \cite{hendrycks19iclr} &  \cmark  & 32.2 & 9.6 & 97.0   \\
Energy \cite{liu20neurips} &  \cmark  & 38.1 & 11.0 & 96.8   \\
GAN negatives \cite{lee18iclr} &  \xmark  & 31.7 & 9.2 & 96.9   \\\hdashline
MOP \cite{dasilva20arxiv} &  \xmark  & 24.5 & 10.9 & 96.0   \\ 
$\text{DPN}^{-}$ \cite{gawlikowski21igrass} &  \cmark  & 27.3 & 9.1 &  97.1  \\
\hdashline
JSDiv (ours) & \cmark   & 38.4   & 12.5  &  96.5   \\
NFlowJS (ours) & \xmark   &  \textbf{44.1} &  \textbf{8.8}  &  \textbf{97.8}  \\\hline
\end{tabular}
\end{footnotesize}
\end{table}

\begin{figure}[ht]
    \centering
    \includegraphics[width=0.88\linewidth]{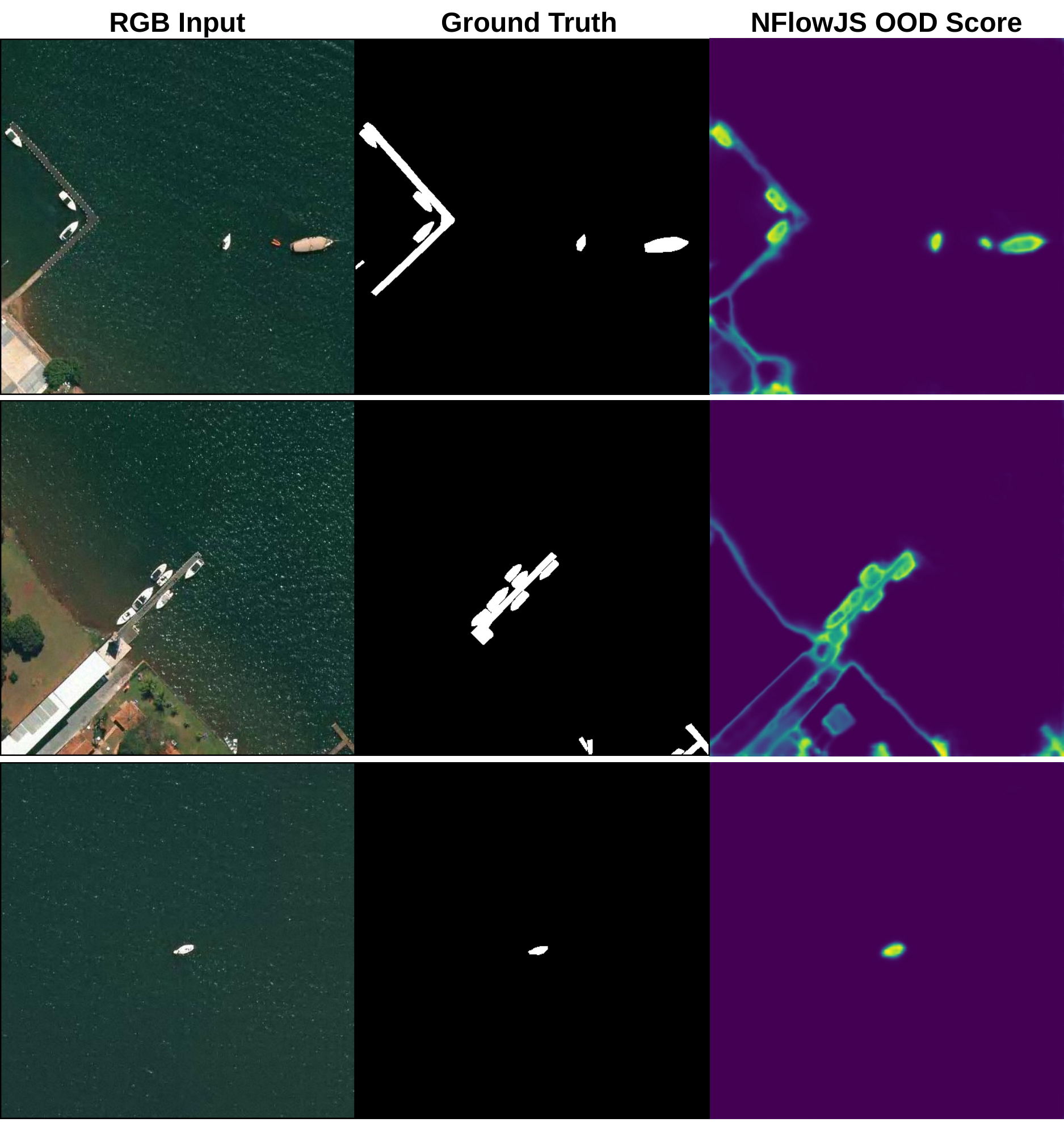}
    \caption{Images from BSB-aerial-OOD (left). Boats and harbour are selected as OOD samples (center). NFlowJS delivers accurate OOD scores (right).
    }
    \label{fig:bsb_fig}
\end{figure}

Figure \ref{fig:bsb_fig} visualizes our performance on the BSB-aerial-OOD dataset.
The left column shows the input images.
The center column shows OOD objects - harbour and boats. 
The right column shows that NFlowJS delivers well-aligned score.

\subsection{Sensitivity of OOD detection to depth}
\begin{figure}[ht]
\centering
\includegraphics[width=.95\linewidth]{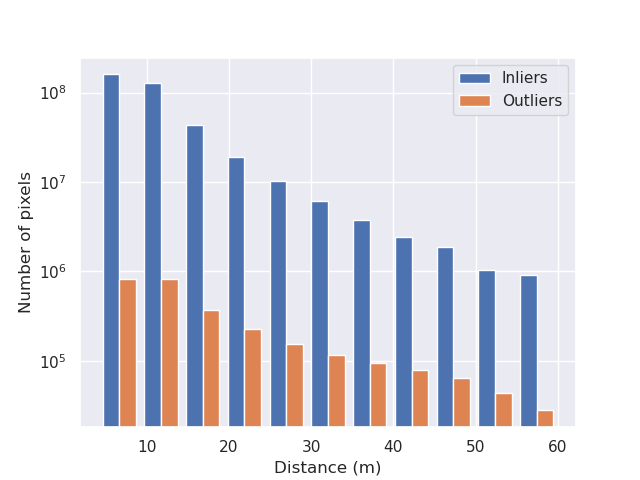}
\caption{Histogram of LostAndFound test-set depth. 62.5\% of anomalous pixels are closer than 15 meters.
We observe significant evaluation bias towards closer ranges.}
\label{fig:distance}
\end{figure}

Self-driving applications challenge us to detect anomalies as soon and as far as possible. 
However, distant anomalies are harder to detect due to being represented with fewer pixels.
We analyze influence of depth to dense OOD detection on the LostAndFound dataset \cite{pinggera16iros}.
The LAF test set consists of 1203 images with the corresponding pixel-level disparity maps and calibration parameters of the stereo rig.
Due to limitations in the available disparity, we perform analysis in the range from 5 to 50 meters.
Figure \ref{fig:distance} shows histograms of inlier and outlier pixels.
More than 60\% of anomalous pixels are closer than 15 meters.
Hence, the usual metrics (AP and $\text{FPR}_{95}$) are biased towards closer ranges.
As we further demonstrate, many methods fail to detect anomalies at larger depths.

We compare our method with the max-logit (ML) and max-softmax \cite{hendrycks17iclr} baselines, ODIN \cite{liang18iclr}, SynBoost \cite{biase21cvpr} and OOD head \cite{bevandic19gcpr}.
Table \ref{tbl:distance_fpr} shows that our method produces low false positive rate even at high distances.
For example, at distances higher than 20 meters we outperform others by a wide margin.
This finding is consistent with Fig.\ \ref{fig:lf-ood} which shows accurate detection of anomalies at larger distances.
\begin{table}[ht]
\caption{Analysis of $\mathrm{FPR}_{95}$ at various distances from the camera.
}
\label{tbl:distance_fpr}
\centering
\begin{footnotesize}
\begin{tabular}{c|cccccc}
\hline \hline
Range & NFlowJS & MSP & ML & SynBoost  & OOD-H  & ODIN \\\hline \hline
5-10 & 0.7 & 16.6 & 4.7 & \textbf{0.2} & 7.9 & 10.9 \\
10-15 & \textbf{1.2} & 18.0 & 7.3 & 17.7 & 10.6 & 9.0 \\
15-20 & \textbf{0.8} & 19.3 & 5.9 & 25.0 & 16.9 & 11.1 \\
20-25 & \textbf{1.1} & 23.2 & 5.8 & 23.3 & 23.6 & 13.4 \\
25-30 & \textbf{1.8} & 28.0 & 7.1 & 18.8 & 26.7 & 16.6 \\
30-35 & \textbf{2.7} & 32.6 & 7.6 & 27.4 & 30.8 & 22.6 \\
35-40 & \textbf{3.5} & 37.9 & 10.1 & 25.4 & 36.8 & 25.9 \\
40-45 & \textbf{5.6} & 41.4 & 13.2 & 25.8 & 42.2 & 30.3 \\
45-50 & \textbf{8.8} & 46.3 & 15.8 & 29.9 & 52.0 & 37.9\\
\hline
\end{tabular}
\end{footnotesize}
\end{table}


\subsection{Inference speed}

A convenient dense OOD detector should not drastically increase already heavy computational burden of semantic segmentation.
Hence, we measure computational overhead of our method and compare it with other approaches.
We measure the inference speed on NVIDIA RTX 3090 for $1024 \times 2048$ inputs.
Table \ref{tbl:infer-speed} shows that SynBoost \cite{biase21cvpr} and SynthCP \cite{xia20eccv} are not applicable for real-time inference due to significant overhead.
The baseline model LDN-121 \cite{kreso21tits} achieves near real-time inference for two megapixel images (46.5 ms, 21.5 FPS).
ODIN \cite{liang18iclr} requires an additional forward-backward pass in order to recover the gradients of the loss with respect to the image.
This results in a 3-fold slow-down with respect to the baseline.
Similarly, MC Dropout \cite{kendall17nips} requires K forward passes for prediction with K MC samples.
This results in 45.8 ms overhead when K=2.
Contrary, our approach increases the inference time for only 7.8 ms with respect to the baseline while outperforming all previous approaches.
The SynthCP measurements are taken from \cite{jung21iccv}.
\begin{table}[ht]
\centering
\caption{Comparison of inference speed on 2MPix images and RTX3090.}
\label{tbl:infer-speed}
\begin{footnotesize}
\begin{tabular}{l|ccc}
\hline\hline
Method & Resynth. & Infer. Time (ms) &  FPS \\ \hline \hline
SynthCP \cite{xia20eccv} &\cmark & 146.9     & 6.8 \\ 
SynBoost \cite{biase21cvpr} & \cmark & 1055.5  & $<1$ \\ \hdashline
LDN-121 (Base) \cite{kreso21tits} & \xmark & 46.5& 21.5 \\ 
Base + ODIN \cite{liang18iclr} &\xmark & +149.1 & 5.11 \\
Base + MC=2 Dropout \cite{kendall17nips} &\xmark & +45.8 & 10.83\\
Base + NFlowJS (ours) &\xmark & +\textbf{7.8} & \textbf{18.4} \\ \hline
\end{tabular}
\end{footnotesize}
\end{table}

\subsection{Visualization of synthetic outliers}

Our method is able to generate samples at multiple resolutions with the same normalizing flow.
The generated samples have a limited variety when compared to a typical negative dataset such as ImageNet or COCO \cite{bevandic19gcpr,chan21iccv}.
Still, training with them greatly reduces overconfidence
 since the model is explicitly trained to produce uncertain predictions in outliers.

\begin{figure}[ht]
    \centering
    \includegraphics[width=0.95\linewidth]{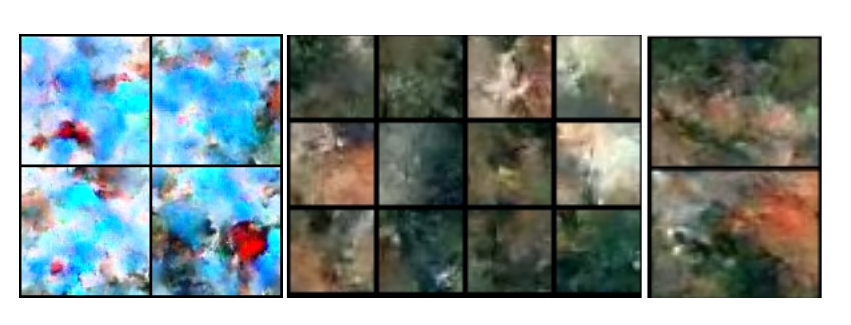}
    \caption{
    Samples of DenseFlow-25-6 after joint training on aerial images.
    }
    \label{fig:outliers-bsb}
\end{figure}

Figure \ref{fig:outliers-bsb} shows synthetic outliers generated by our normalizing flow after joint training on aerial images.

\begin{figure}[ht]
    \centering
    \includegraphics[width=0.85\linewidth]{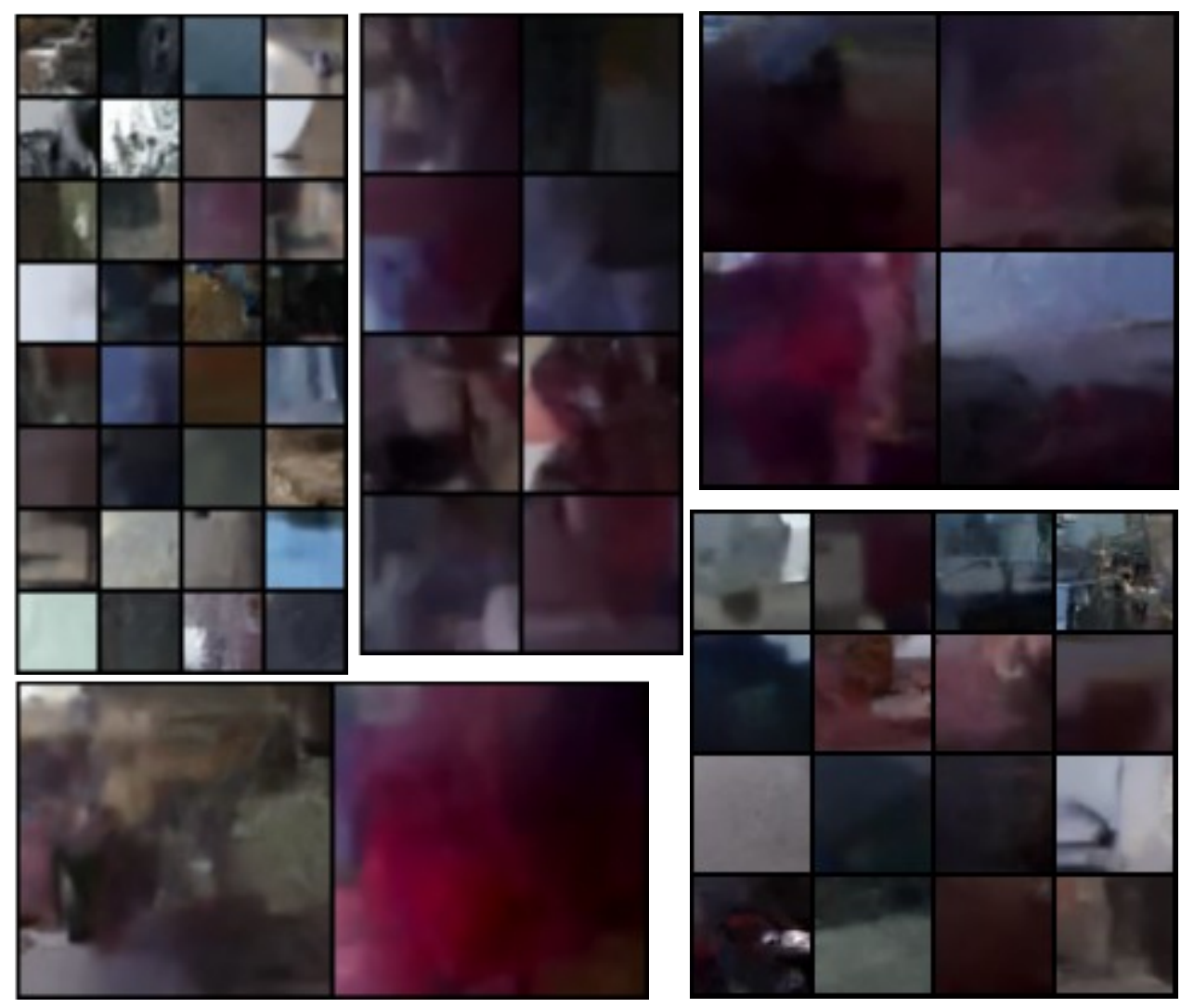}
    \caption{
    Samples of DenseFlow-25-6 after joint training on road-driving images as proposed in section \ref{sec:neg_ds_to_flow}.
    }
    \label{fig:outliers-smiyc}
\end{figure}

Figure \ref{fig:outliers-smiyc} shows samples of a normalizing flow after joint training on road-driving scenes.
Comparison with Figure \ref{fig:outliers-bsb} reveals that the appearance of our synthetic negative samples
strongly depends on the underlying inlier dataset.
Samples from Figure \ref{fig:outliers-bsb} resemble lakes and forests
while samples from Figure \ref{fig:outliers-smiyc} resemble 
road, sky, cars and buildings.
These observations do not come up as a surprise
since our normalizing flows are trained
to generate data points along the border
of the inlier distribution (cf.\ Fig. \ref{fig:toy_nfjs}).
In other words, our method patches the open-space risk 
of a particular segmentation model
by adapting the synthetic negative data
to the training dataset.

\subsection{Synthetic negatives and the separation in the feature space}

Up to now, we have considered softmax-activated models.
However, softmax can assign arbitrarily large probabilities regardless of the distance from the closest training datum in the feature space \cite{scheirer14tpami}, which fails to bound open-space risk \cite{scheirer12tpami,boult19aaai}.
We analyze the usefulness of synthetic negatives in open-set recognition by considering a popular baseline that is denoted as max-logit \cite{hendrycks19arxiv,vaze22iclr,chen22tpami}.
Max-logit value is proportional to  the projection of the feature vector of a given sample onto the closest class prototype vector.
This value can be thresholded to bound the open space risk \cite{boult19aaai,stefano00tsmc}.

\begin{figure}[ht]
    \centering
    \includegraphics[width=0.95\linewidth]{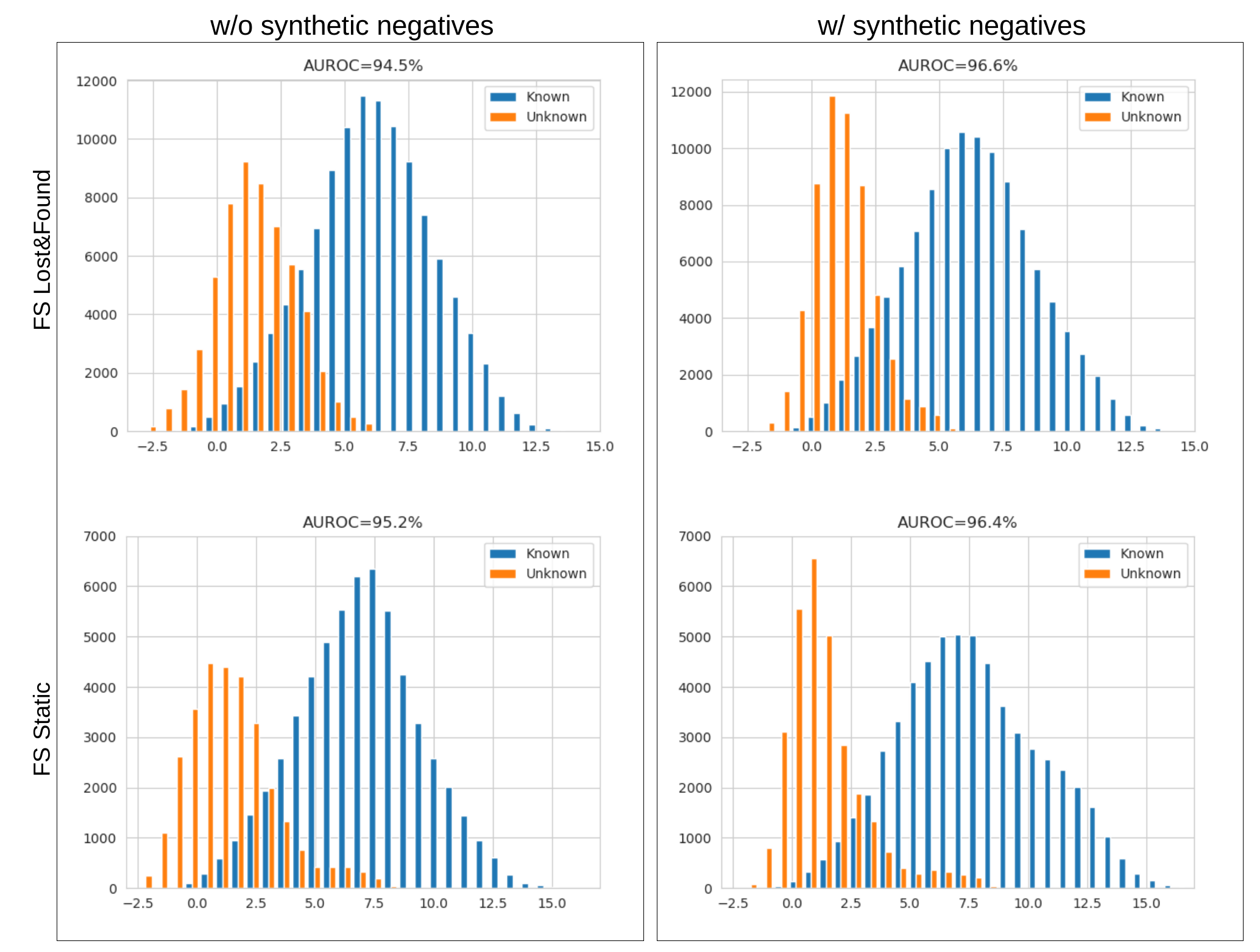}
    \caption{
    Training on synthetic negative data improves the separation between test inliers and test outliers in the feature space.}
    \label{fig:syn_ft_ml}
\end{figure}

The left column of Figure \ref{fig:syn_ft_ml} shows histograms of max-logit values for known and unknown pixels on Fishyscapes val.
The right column shows the same histograms after fine-tuning with our synthetic negative samples.
The figure shows that training with our synthetic negatives
increases the separation between 
known and unknown pixels in feature space, 
and improves the AUROC score.
Similar effects have been reported
after training with real negative data \cite{neal18eccv,kong22tpami}.
However, as argued before, our approach avoids
the bias towards test anomalies 
that are related to the training data.
Furthermore, it offers a great alternative
for non-standard domains as shown in Table \ref{tbl:bsb_ood}. 
Hence, the proposed method appears as
a promising component of future approaches for dense open-set recognition.

\section{Ablations}

We ablate the impact of loss in negative pixels, the choice of generative model, the impact of pretraining, as well as the impact of temperature scaling on dense OOD detection.

\subsection{Impacts of the loss function and OOD score}
\label{sec:abl}
Table \ref{tbl:abl-loss} analyzes the impact of the loss function $L_{neg}$ and the OOD score $s_\delta$ on AnomalyTrack val and ObstacleTrack val. 
The two chosen datasets feature large and small anomalies, respectively.
We separately validate the modulation factor $\lambda$ for each choice of the negative loss, as well as the temperature parameter.
We use T=10 for max-softmax and T=2 for divergence-based scoring functions.
We report average performance over last three epochs.
Row 1 shows the standard setting with KL divergence as $L_{\mathrm{neg}}$ and max-softmax as the OOD score \cite{lee18iclr,hendrycks19iclr}.
Row 2 uses KL divergence both as the loss function and the OOD score.
Row 3 features the reverse KL divergence. 
Minimizing the reverse divergence between the uniform distribution and the softmax distribution is equivalent to maximizing the softmax entropy \cite{chan21iccv}.
Rows 4 and 5 feature the JS divergence.
We observe that the reversed KL divergence outperforms the standard KL-MSP setup in 3 out of 4 metrics.
However, JS divergence substantially outperforms all alternatives both as the loss function (JSD-MSP vs KL-MSP) and as the OOD score (JSD-JSD vs JSD-MSP and RKL-RKL).
We explain this advantage with robust response in synthetic outliers which resemble inliers, as well as with improved consistency during training and scoring (cf.\ sections \ref{sec:div-loss} and \ref{sec:div-inf}).

\begin{table}[ht]
\centering
\caption{Validation of the loss in negative pixels and the OOD score.
}
\label{tbl:abl-loss}
\begin{footnotesize}
\begin{tabular}{ll|cccc}
\hline \hline
 \multirow{2}{*}{Loss}& \multirow{2}{*}{$s(\mathbf{x})$} & \multicolumn{2}{c|}{AnomalyTrack-val} & \multicolumn{2}{c}{ObstacleTrack-val} \\
 &  & AP & \multicolumn{1}{c|}{$\mathrm{FPR}_{95}$} & AP & \multicolumn{1}{c}{$\mathrm{FPR}_{95}$}\\ \hline \hline
 KL & MSP & 57.5 $\pm$ 0.7& 29.0 $\pm$ 1.7& 95.1 $\pm$ 0.2& 0.2 $\pm$ 0.1\\
 KL & KL & 55.7 $\pm$ 0.4& 26.3 $\pm$ 1.3& 94.3 $\pm$ 0.2& 0.1 $\pm$ 0.0\\
 RKL & RKL & 57.0 $\pm$ 0.3& 28.9 $\pm$ 0.3& 94.4 $\pm$ 0.1 & 0.3 $\pm$ 0.0\\
 JSD & MSP & 63.0 $\pm$ 0.5& 22.8 $\pm$ 0.7 & \textbf{96.1}  $\pm$ 0.2 & 0.2  $\pm$ 0.0\\
 JSD & JSD & \textbf{63.3} $\pm$ 0.6& \textbf{19.8} $\pm$ 0.8& 95.8 $\pm$ 0.2& \textbf{0.1} $\pm$ 0.0\\
\hline
\end{tabular}
\end{footnotesize}
\end{table}

\subsection{Impact of the choice of generative model}

Table \ref{tbl:abl-gm} compares synthetic negative data generated by normalizing flow with synthetic negative data generated by GAN\cite{lee18iclr} and synthetic negative pre-logit features generated by GMM \cite{du22iclr}.
Interestingly, training on synthetic OOD features produced by GMM achieves better average precision than synthetic negative images generated by GAN.
Still, generating synthetic negatives with a normalizing flow outperforms both GAN images and GMM features.
This advocates for the advantages of maximum likelihood over adversarial training for the generation of synthetic negatives, as described in  Sec.\ \ref{sec:theoretical_analysis}.

\begin{table}[ht]
\centering
\caption{
Impact of generative model to OOD detection performance.
}
\label{tbl:abl-gm}
\begin{footnotesize}
\begin{tabular}{c|cccccc}
\hline \hline
 \multirow{2}{*}{Generator} & \multicolumn{2}{c|}{AnomalyTrack-val} & \multicolumn{2}{c}{ObstacleTrack-val}\\
   &  \multicolumn{1}{c}{AP} & \multicolumn{1}{c|}{$\mathrm{FPR}_{95}$} & AP & \multicolumn{1}{c}{$\mathrm{FPR}_{95}$}\\ \hline \hline
    GMM-VOS & 56.7 $\pm$ 0.2 & 28.0 $\pm$ 0.4 & 81.8 $\pm$ 0.5 & 3.9 $\pm$ 0.2 \\
 GAN  & 56.1 $\pm$ 0.4 & 26.1 $\pm$ 0.6 & 80.8 $\pm$ 0.4 & 3.6 $\pm$ 0.1 \\
  NFlow & \textbf{61.4} $\pm$ 0.8& \textbf{21.7} $\pm$ 1.3& \textbf{94.9} $\pm$ 0.1& \textbf{0.1} $\pm$ 0.1\\
\hline
\end{tabular}
\end{footnotesize}
\end{table}

\subsection{Impact of pre-training}

Table \ref{tbl:abl-train} explores the impact of pre-training to OOD detection performance.
Row 1 shows the performance when neither generative nor discriminative model are trained prior to the joint training (Section \ref{sec:neg_ds_to_flow}).
In this case, we jointly train both models from their random initializations.
Row 2 reveals that discriminative pre-training improves OOD detection.
Introducing the synthetic negatives after discriminative pre-training improves generalization.
Row 3 shows that pre-training both models generalizes even better.
\begin{table}[ht]
\centering
\caption{Impact of pre-training to the success of joint training.
}
\label{tbl:abl-train}
\begin{footnotesize}
\begin{tabular}{cc|cccccc}
\hline \hline
 \multirow{2}{*}{Cls.} & \multirow{2}{*}{Flow} & \multicolumn{2}{c|}{AnomalyTrack-val} & \multicolumn{2}{c}{ObstacleTrack-val}\\
   & &  \multicolumn{1}{c}{AP} & \multicolumn{1}{c|}{$\mathrm{FPR}_{95}$} & AP & \multicolumn{1}{c}{$\mathrm{FPR}_{95}$}\\ \hline \hline
 \xmark & \xmark & 56.9 $\pm$ 1.2& 27.8 $\pm$ 2.1& 90.5  $\pm$ 0.3& 1.0 $\pm$ 0.1\\
 \cmark & \xmark & 61.4 $\pm$ 0.8& 21.7 $\pm$ 1.3& 94.9 $\pm$ 0.1& 0.1 $\pm$ 0.1\\
 \cmark & \cmark & \textbf{63.3} $\pm$ 0.6& \textbf{19.8} $\pm$ 0.8& \textbf{95.8} $\pm$ 0.2& \textbf{0.1} $\pm$ 0.0\\
\hline
\end{tabular}
\end{footnotesize}
\end{table}

\subsection{Impact of temperature scaling}
Table \ref{tbl:abl-temp} shows the impact of softmax recalibration to OOD detection.
The table explores three different temperatures.
We observe that temperature scaling significantly improves Jensen-Shannon scoring.
We also note that utilizing RealNVP \cite{dinh17iclr} instead of DenseFlow \cite{grcic21neurips} decreases OOD detection performance.

\begin{table}[ht]
\centering
\caption{Impact of temperature to JSD scoring.}
\label{tbl:abl-temp}
\begin{footnotesize}
\begin{tabular}{l|cccccc}
\hline \hline
 \multirow{2}{*}{Temp.} & \multicolumn{2}{c|}{AnomalyTrack-val} & \multicolumn{2}{c}{ObstacleTrack-val}\\
   &  \multicolumn{1}{c}{AP} & \multicolumn{1}{c|}{$\mathrm{FPR}_{95}$} & AP & \multicolumn{1}{c}{$\mathrm{FPR}_{95}$}\\ \hline \hline
 T=1 & 59.7 $\pm$ 0.5& 40.0 $\pm$ 0.8& 92.6  $\pm$ 0.3& 1.1 $\pm$ 0.1\\
 T=1.5 & 62.7 $\pm$ 0.6& 23.7 $\pm$ 0.9&95.3 $\pm$ 0.2& 0.2 $\pm$ 0.0\\
 T=2 & \textbf{63.3} $\pm$ 0.6& \textbf{19.8} $\pm$ 0.8& \textbf{95.8} $\pm$ 0.2& \textbf{0.1} $\pm$ 0.0\\
\hline
\end{tabular}
\end{footnotesize}
\end{table}

\section{Conclusion}
We have presented a novel method for dense OOD detection and  outlier-aware semantic segmentation. 
Our method trains on mixed-content images obtained by pasting synthetic negative patches into training images.
We produce synthetic negatives 
by sampling a generative model 
which is  jointly trained
to maximize the likelihood
and to give rise to uniform predictions at the far end of the discriminative model.
Such collaborative learning leads to conservative outlier-aware predictions which are suitable for OOD detection and  outlier-aware semantic segmentation.

We extend the previous work with the following consolidated contributions.
First, we replace the adversarial generative model (GAN) with a normalizing flow.
We believe that the resulting improvement is due to better coverage of the training distribution.
Second, we extend the collaborative training setup for dense prediction.
Generative flows are especially well-suited for this task due to straightforward generation at different resolutions.
Third, we improve the performance by pre-training the normalizing flow and the discriminative model 
prior to joint training. 
Fourth, we propose to use JS divergence as a robust criterion for training a discriminative model with synthetic negatives.
We also show that the same criterion can be used as a principled and competitive replacement for ad-hoc scoring functions such as max-softmax. 

We have evaluated the proposed method on standard benchmarks and datasets for dense OOD detection and outlier-aware segmentation.
The results indicate a significant advantage with respect to all previous approaches on the majority of the datasets from two different domains.
The advantage becomes substantial 
in the case of non-standard domains
with few suitable auxiliary datasets
for sampling real negative data.
Additionally, we demonstrate a great potential of our method for real-world deployments due to minimal computational overhead.
Suitable avenues for future work include 
extending our method to setups with bounded open-set risk and other dense prediction tasks.

\newpage
\section*{Acknowledgements}
The authors thank Jakob Verbeek for suggesting normalizing flows as a promising tool for generating synthetic negative samples. This work has been supported by Croatian Science Foundation (grant IP-2020-02-5851 ADEPT), European Regional Development Fund (grant KK.01.2.1.02.0119 A-Unit, grant KK.01.1.1.01.0009 DATACROSS).

\bibliographystyle{IEEEtran}
\bibliography{main}

\section{Biography Section}
 




\begin{IEEEbiographynophoto}{Matej Grcić} received a M.Sc. degree from the Faculty of Electrical Engineering and Computing in Zagreb. He finished the master study program in
Computer Science in 2020. He is pursuing his Ph.D. degree at Uni-ZG FER.
His research interests include generative modeling and open-world recognition.
\end{IEEEbiographynophoto}
\begin{IEEEbiographynophoto}{Petra Bevandić} received a M.Sc. degree from the Faculty of Electrical Engineering and Computing in Zagreb. She finished the master study
program in Computer Science in 2014. She is pursuing her Ph.D. degree at
Uni-ZG FER. Her research interests include robust and open-world recognition.
\end{IEEEbiographynophoto}

\begin{IEEEbiographynophoto}{Zoran Kalafatić} received a Ph.D. degree in computer science from the
University of Zagreb, Croatia. He is currently an associate professor at Uni-ZG
FER. His research interests focus on deep convolutional architectures for
classification, object detection and tracking.
\end{IEEEbiographynophoto}

\begin{IEEEbiographynophoto}{Siniša Šegvić} received a Ph.D. degree in computer science from the
University of Zagreb, Croatia. He was a post-doctoral researcher at IRISA
Rennes and also at TU Graz. He is currently a full professor at Uni-ZG
FER. His research interests focus on deep convolutional architectures for
classification and dense prediction.
\end{IEEEbiographynophoto}

\vfill

\end{document}